\documentclass[11pt, copyright, logo, secondlogo]{googledeepmind}
\renewcommand{\today}{}  

\usepackage{amssymb}
\usepackage{amsmath}
\usepackage{amsthm}
\usepackage{algorithm}
\usepackage{algpseudocode}
\usepackage{xspace}
\usepackage[authoryear, sort&compress, round]{natbib}
\usepackage{bbm}
\usepackage{wrapfig}
\usepackage{xcolor}
\usepackage[multiple]{footmisc}

\usepackage{verbatim}

\newcount\Comments  %
\newcommand{\kibitz}[2]{\ifnum\Comments=1\textcolor{#1}{#2}\fi}
\definecolor{red}{rgb}{1,0,0}
\definecolor{darkred}{rgb}{0.5,0,0}
\definecolor{orange}{rgb}{1.0,0.64,0}
\definecolor{darkgreen}{rgb}{0,0.5,0}
\definecolor{darkblue}{rgb}{0,0,0.7}
\definecolor{purple}{rgb}{.6, 0,.6}

\newcommand{\cpp}{\mbox{C\texttt{++}}\xspace} 
\newcommand{\noise}{\varepsilon}

\newcommand{\x}{\ensuremath{x}}
\newcommand{\y}{\ensuremath{y}}
\newcommand{\abs}[1]{\lvert #1 \rvert}

\newcommand{\ymedian}{\ensuremath{y_{\operatorname{median}}}}

\newcommand{\scaledx}{\widehat{\x}}
\newcommand{\scaledy}{\widehat{\y}}
\newcommand{\vecx}{\scaledx} 
\newcommand{\vecy}{\scaledy} 

\newcommand{\poolsize}{\ensuremath{P}}
\newcommand{\flybatch}{\ensuremath{\mathbf{\widetilde{X}}}}
\newcommand{\flybatchsize}{\ensuremath{p}}

\newcommand{\barf}{\overline{f}}
\newcommand{\veclambdalog}{\overrightarrow{\lambda}_{\log}}
\newcommand{\R}{\mathbb{R}}
\newcommand{\alg}{\ensuremath{\mathcal{A}}}
\newcommand{\X}{\mathcal{X}}
\newcommand{\Y}{\mathcal{Y}}
\newcommand{\HV}{\mathcal{HV}}
\newcommand{\norm}[1]{\left\lVert #1 \right\rVert}
\DeclareMathOperator*{\argmax}{argmax}

\newcommand{\rbudget}[1]{\ensuremath{\operatorname{RequiredBudget}\paren{#1}}}

\newcommand{\eratio}[1]{\ensuremath{\operatorname{EfficiencyRatio}\paren{#1}}}

\newcommand{\paren}[1]{\ensuremath{\left(#1\right)}}
\newcommand{\set}[1]{\ensuremath{\left\{#1\right\}}}
\newcommand{\expct}[1]{\ensuremath{\mathbb{E}\text{$\left[#1\right]$}}}
\newcommand{\floor}[1]{\ensuremath{\left \lfloor {#1} \right \rfloor }}
\newtheorem{theorem}{Theorem}
\newtheorem{definition}{Definition}

\title{The Vizier Gaussian Process Bandit Algorithm}

\author[1]{Xingyou Song}
\author[1]{Qiuyi Zhang}
\author[1]{Chansoo Lee}
\author[2]{Emily Fertig}
\author[1]{Tzu-Kuo Huang}
\author[1]{Lior Belenki}
\author[1]{Greg Kochanski}
\author[1]{Setareh Ariafar}
\author[2]{Srinivas Vasudevan}
\author[1]{Sagi Perel}
\author[1]{Daniel Golovin}

\affil[1]{Google DeepMind}
\affil[2]{Google Research}

\begin{abstract}
Google Vizier has performed millions of optimizations and accelerated numerous research and production systems at Google, demonstrating the success of Bayesian optimization as a large-scale service. Over multiple years, its algorithm has been improved considerably, through the collective experiences of numerous research efforts and user feedback. In this technical report, we discuss the implementation details and design choices of the current default algorithm provided by Open Source Vizier. Our experiments on standardized benchmarks reveal its robustness and versatility against well-established industry baselines on multiple practical modes.
\end{abstract}

\begin{document}

\maketitle

\section{Introduction}
Google Vizier \citep{kdd_vizier} is one of the world's largest black-box optimization services; it has tuned more than 70 million objectives across Google, and is also available to the public via Vertex Vizier, a Google Cloud product. 
 
Vizier's default algorithm is loosely based on Gaussian process bandit optimization \citep{gp_bandit}, but has evolved over time to improve not only standard performance metrics such as regret and optimality gap, but also user experience, inference speed, flexibility, scalability, and reliability.

\vspace{0.3cm}
\begin{figure}[h]
    \centering
    \includegraphics[width=1.0\textwidth]{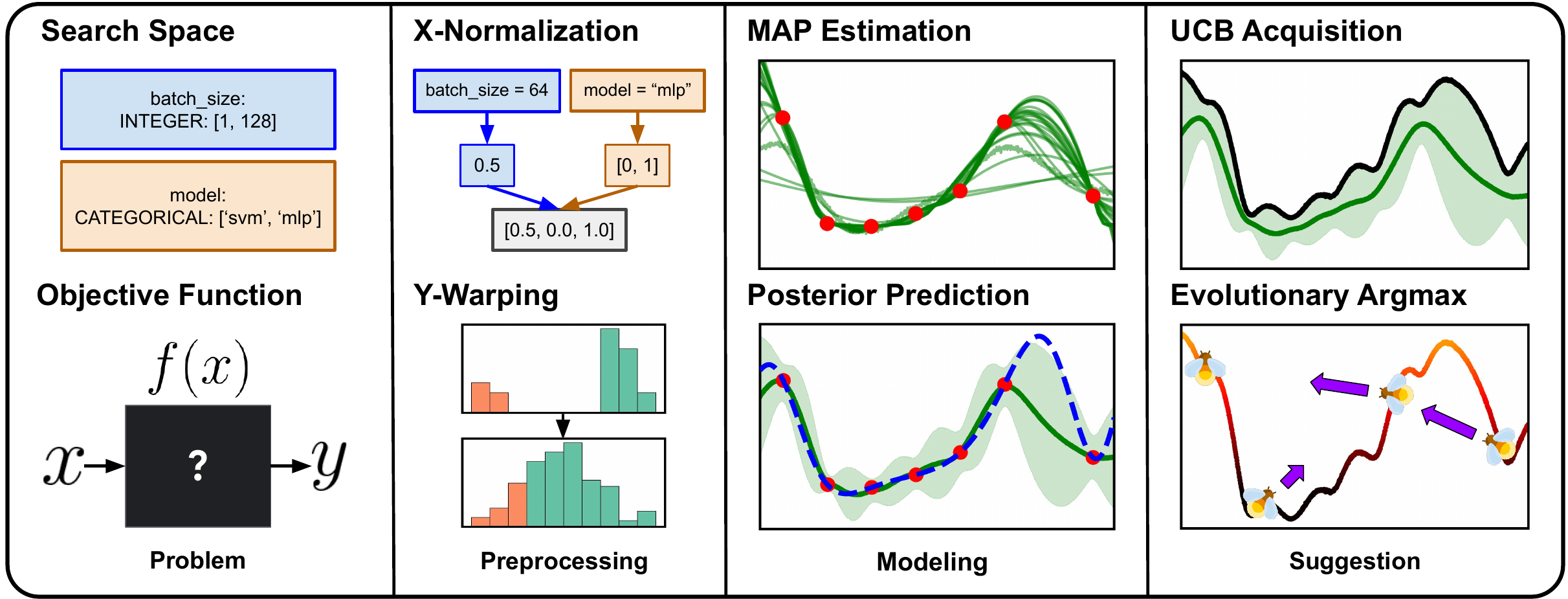}
    \vspace{0.04cm}
    \caption{Key components of the Google Vizier Bayesian optimization algorithm.\vspace{0.2cm}}
\end{figure}

As Google Vizier predates the advent of autodifferentiation packages such as TensorFlow \citep{tensorflow} and JAX \citep{jax} and even the emergent popularity of Python, many design choices were based on the programming philosophy of \cpp, whose advantages over Python include full control over all low-level details and extensive multithreading support. Additionally, the algorithm was supplemented by much of the low-latency backend service logic for handling requests, all of which were also written in \cpp to support internal Google users and Vertex AI users.

However, support for Python has risen dramatically, due to its flexibility and ease of implementation, not to mention the numerous auto-differentiation packages such as TF Probability \citep{tfp} which now perform computations such as Gaussian Process (GP) regression orders of magnitude faster via GPU accelerators. Furthermore, the recent release of Open Source Vizier \citep{oss_vizier} allows a customizable backend service written entirely in Python, available for the wider research community. 

While prior work \citep{kdd_vizier, oss_vizier} discussed the overall Vizier system, in this paper, we provide a reference on the Vizier default algorithm and include many innovations since then. Our implementation is a numerically stable and production-grade Python reproduction. Our specific contributions are as follows:
\begin{itemize}
\item We formalize the current version of the Google Vizier default algorithm and explain its features, design choices and lessons learned throughout its iteration, which will be of use to both researchers and practitioners alike. 
\item We provide an open source Python implementation, powered by TF Probability on JAX, of the original Google \cpp based Vizier algorithm.
\item We benchmark Vizier against industry-wide baselines and demonstrate its competitive robustness on multiple axes, including high-dimensional, categorical, batched, and multi-objective optimization. 
\item We further ablate the unconventional design choice of using a zeroth-order evolutionary acquisition optimizer and discuss its key strengths.
\end{itemize}

\textbf{Disclaimer:} While additional scenarios such as contextual bandit optimization, transfer learning, and conditional parameter search are supported in Google Vizier, we do not discuss their algorithmic approaches in this open-source focused paper. 

\section{Preliminaries}
\subsection{Blackbox Optimization}
We use the standard terminology found in \citep{kdd_vizier} and \citep{oss_vizier}, where an optimization \textit{study} consists of a \textit{problem statement} and trajectory of \emph{trials}. A problem statement consists of: 

\begin{itemize}
\item Search space $\X$, which is the cartesian product of parameters of type \texttt{DOUBLE}, \texttt{INTEGER}, \texttt{DISCRETE}, or \texttt{CATEGORICAL}. Except for \texttt{CATEGORICAL} types, each parameter also can have an associated scaling type (linear, log, or reverse log). 
\item Measurement space $\mathcal{Y} \subseteq \mathbb{R}^M$ which consists of $M$ \emph{metrics} from the blackbox objective function.
\end{itemize}
The blackbox function to be optimized $f: \X \rightarrow \mathcal{Y}$ is unknown to the algorithm, and may be stochastic. A \textit{trial} consists of a pair $(x,y)$, where $x \in \X$ is the suggested parameters for the blackbox objective function, and $y = f(x) \in \mathcal{Y}$ is a \emph{measurement}. Assuming a canonical ordering of parameters and metrics within a study, we denote $x^{(i)}$ and $y^{(i)}$ to be the $i$-th parameter and its associated measurement, respectively. To simplify notation, we assume a single-metric objective function $f: \X \rightarrow \R$ unless otherwise specified.

A trajectory of trials $(x_{1}, y_{1}, x_{2}, y_{2}, ..., x_{t}, y_{t})$ can be obtained by an iterative optimization loop where an algorithm proposes new parameters $x_{t}$ and obtains  $y_{t} = f(x_{t})$ through (possibly noisy) evaluation. Abstractly, the goal of blackbox optimization is to quickly improve $y$ through multiple iterations which can be concretely defined with a performance measure over the trajectory such as best-so-far (corresponding to simple regret) or average performance (corresponding to cumulative regret), based on some fixed budget on the number of trials $T$. In this paper, we assume the goal is to maximize the objective, hence we seek trajectories $(x_t, y_t)_{t \ge 0}$ with the highest possible value of $\max_{t} \set{f(x_t)}$.

\subsection{Bayesian Optimization}

We provide a brief summarization of the GP-UCB algorithm \citep{gp_bandit} for Bayesian optimization. Given a Gaussian process with a prior mean $\mu_0 : \X \rightarrow \R$ and kernel $K: \X \times \X \rightarrow \R_{>0}$, we can define its mean and standard deviation functions respectively as $\mu: \X \rightarrow \R$ and $\sigma: \X \rightarrow \R_{>0}$.

\vspace{0.2cm}
\begin{algorithm}[h]
\caption{Gaussian Process Upper Confidence Bound (GP-UCB) with parameter $\beta > 0$}
\begin{algorithmic}
\For{$t \in \{1, 2, ... \}$}
\State {Compute posteriors to obtain $\mu_{t}(x)$ and $\sigma_{t}(x)$}
\State {$x_{t} = \arg \max_{x \in \X} \mu_{t}(x) + \sqrt{\beta} \cdot \sigma_{t}(x)$}
\State {Evaluate $y_{t} = f(x_{t})$} 
\EndFor
\end{algorithmic}
\label{alg:gp_ucb}
\end{algorithm}
\vspace{0.2cm}

While this algorithm is conceptually clear, there are many details which crucially affect overall optimization performance and algorithm serving in a distributed production system.

\begin{itemize}

\item \textbf{Input Preprocessing:} Transforming the parameters into a unit $L$-infinity ball eases modeling.

\item \textbf{Output Preprocessing:} Nonlinear transformations of the measurement $y \in \R^{M}$ into a smaller subspace can mitigate harmful effects of outliers and shift model capacity towards discerning finer distinctions among good trials and away from poor trials.

\item \textbf{Model Choice:} The choice of \emph{kernel} and priors over its hyperparameters are critical~\citep{prior_gp}. For posterior updates, we use the Maximum A Posteriori (MAP) estimates of the kernel hyperparameters.

\item \textbf{Acquisition Function:} UCB is but one option of many acquisition functions which influence explore / exploit tradeoffs, diversity of selected trials, and handling of multiple metrics.

\item \textbf{Acquisition Maximization:} In general, maximizing the acquisition function is non--trivial, and the acquisition optimizer affects overall performance significantly.

\end{itemize}

\section{Vizier's Gaussian Process Bandits}
In this section, we outline the key components of the default Vizier algorithm, with additional details in Appendix \ref{appendix:vizier_gp_bandit_extended}. Over time, these components on the C++ stack co-evolved, and thus form something approximating a local optimum in the design space.

\subsection{Preprocessing}
In order to improve the accuracy of our Gaussian Process regression model, we preprocess both inputs and outputs into $\scaledx$ and $\scaledy$ respectively, which are also reversible back into the original $\X$, $\Y$ spaces. \textbf{Note:} For conciseness in the paper, we use $\scaledx, \scaledy$ notation only when needed for mathematical precision.
\subsubsection{Input Preprocessing} To provide reasonable scales as inputs, we enforce $\scaledx \in [0, 1]^{D}$ for non-\texttt{CATEGORICAL} parameters. Using the provided bounds $x^{(d)}_{\min}$ and $x^{(d)}_{\max}$, for linear scaling, we transform the original value $x^{(d)}$ into $(x^{(d)} - x^{(d)}_{\min})/(x^{(d)}_{\max}  - x^{(d)}_{\min})$. For log scaling, we first apply $\log(\cdot)$ to the input and bounds before applying the scaling transformation, and apply analogously for reverse-log scaling. 

\subsubsection{Output Preprocessing} 
\begin{figure}[t]
    \centering
    \includegraphics[width=\textwidth]{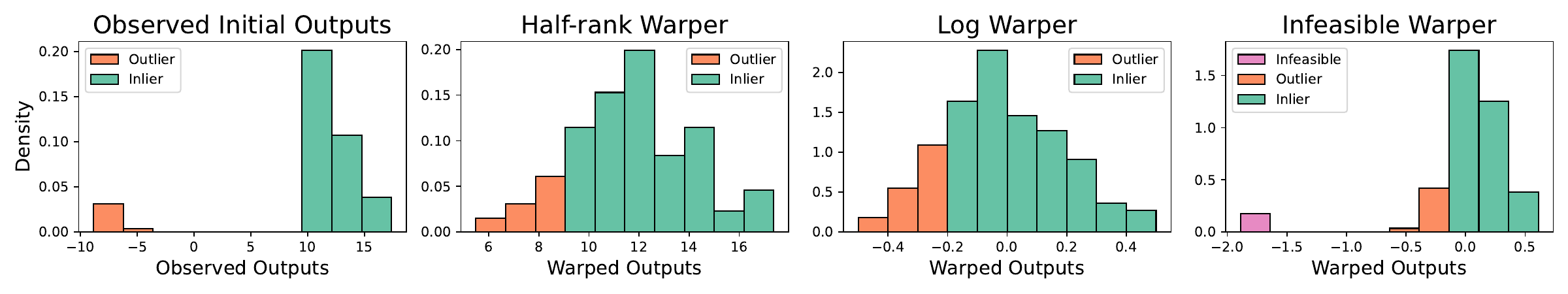}
    \caption{\small Illustration of warped objectives through the pipeline. Initially, objectives form a non-Gaussian distribution with a few outliers (infeasible metrics not shown). Plots display the outcome distribution of applying each warper in order. Note: After the infeasible warper, infeasible metrics are converted to feasible values.}
    \label{fig:output-warper}
\end{figure}

To provide $y$ objectives as suitable formats $\scaledy$ for GPs, we perform multiple stages of transformations, all of which are \textit{relative} to the observed values $(y_1, y_2, ..., y_t)$.
These are asymmetric and assume maximization is the goal.
Important \textit{warping} effects are displayed in Fig. \ref{fig:output-warper}, and the entire pipeline consists of, in order:
\begin{itemize}


\item \textbf{Linear Scaling:} Let $\xi(I) := \sqrt{\frac{1}{|I|} \sum_{i \in I}(\y_i-\ymedian)^2}$. Let our deviation be $\xi(\set{i : \y_i \ge \ymedian})$ if non--zero, else by $\xi(\set{i : 1 \le i \le t})$ if non--zero. Shift the y-values so the median is zero, and then divide by the deviation.

\item \textbf{Half-Rank Warping:} Supresses harm from extremely bad outliers. This warping rescales all unpromising objectives (i.e., worse than the median) so that scaled values are distributed as the lower half of a normal distribution, $-|\mathcal{N}(0, 1)|$, ensuring the typical deviation for poor objectives roughly matches those for good objectives.
Median or better objectives remain unchanged.

\item \textbf{Log Warping:} Increases modeling resolution for good values. We first apply the normalization $\scaledy \leftarrow \frac{y_{max} - y}{y_{max} - y_{min}}$ and then the warping $\scaledy \leftarrow 0.5 - \log(1+(s-1) \cdot \scaledy) / {\log(s)}$ where $s$ is a free parameter set to $1.5$. 
This gently stretches the intervals between better y-values and compresses the intervals between worse values.

\item \textbf{Infeasibility Warping:} We replace all infeasible objectives with $y_{min} - 0.5 (y_{max} - y_{min}) $. This maps infeasible objectives to a reasonably scaled unpromising value to prevent the optimizer from re-exploring infeasible regions.

\item \textbf{Mean Shifting:} Finally, we shift the objectives to have zero mean.

\end{itemize}

\subsection{Gaussian Process Model}

%

User objectives are commonly Lipschitz-continuous over preprocessed featurized search spaces, with respect to e.g., coordinated-scaled Euclidean distance. Thus we use a Matern-5/2 kernel \citep{gp_book} along with length-scaled automatic relevance determination (ARD) in the GP model. The probabilistic model is defined as follows (details in Appendix \ref{appendix:vizier_gp_bandit_extended}), where $\mathcal{N}_{[a,b]}$ is the normal distribution truncated to range $[a,b]$, parameterized by mean and variance:

\begin{itemize}
\item $\alpha_{\log} \sim \mathcal{N}_{[-3, 1]}(\log 0.039, 50)$ is the log of the amplitude of Matern kernel.
\item $\lambda_{\log}^{(d)} \sim \mathcal{N}_{[-2, 1]}(\log 0.5, 50)$ is the log of the squared length scale for the $d$-th dimension, i.i.d. for $d=1,\ldots,D$. Denote $\veclambdalog = (\lambda_{\log}^{(1)}, \ldots, \lambda_{\log}^{(D)}).$
\item $\noise_{\log} \sim \mathcal{N}_{[-10, 0]}(\log 0.0039, 50)$ is the log of the standard deviation of the Gaussian noise.
\item $K(\cdot, \cdot)$ is the Matern-5/2 kernel with amplitude $\exp(\alpha_{\log})$ and length scales $\sqrt{\exp(\veclambdalog)}$, applied over $\scaledx$.
\item $\barf \sim \mathcal{GP}(0, K)$ is an objective function in post-processed space sampled from a Gaussian Process with zero-mean and kernel $K$.
\item $\scaledy \sim \mathcal{N}\left(\barf(\scaledx), \exp(\noise_{\log}) \right)$ is a noisy objective prediction in post-processed space, given $\scaledx$.
\end{itemize}

We call $\alpha_{\log}, \veclambdalog$ and $\noise_{\log}$ \emph{kernel hyperparameters}. Note that we define priors over the \emph{log} of kernel hyperparameters in order to avoid the extra $1/x$ term in the pdf of the log normal distributions. We use a relatively narrow truncation range for the kernel hyperparameters \citep{prior_gp} and a zero mean function. This is possible because we model preprocessed input and ouptut values.

Note that we do not one-hot embed \texttt{CATEGORICAL} parameters, as this leads to large uncertainties in unsupported regions. For example, a \texttt{CATEGORICAL} parameter with feasible values \texttt{\{``foo'', ``bar''\}} will be one-hot embedded into $\R^2$, but support points are only $(0, 1)$ or $(1, 0)$. Similarly to other open-source packages such as Ax \citep{botorch_official} and Optuna \citep{optuna}, we calculate distance contributions from \texttt{CATEGORICAL} dimensions to the Matern-5/2 kernel separately. Each \texttt{CATEGORICAL} parameter $x^{(c)}$ produces a distance contribution $\mathbbm{1}\paren{x^{(c)}_{i} \neq x^{(c)}_{j}}$ and uses a single length scale across its categories.

\subsection{Posterior Updates}

Given observed data $\{\scaledx_{s}, \scaledy_{s}\}_{s=1}^{t}$, we seek the MAP estimate of kernel hyperparameters which maximizes the joint probability $\mathrm{Pr}(\{\scaledx_{s}, \scaledy_{s}\}_{s=1}^{t},  \alpha_{\log}, \veclambdalog, \noise_{\log})$. That is, we maximize \begin{equation}\log \mathrm{Pr}(\alpha_{\log}) + \log \mathrm{Pr}(\veclambdalog) + \log \mathrm{Pr}(\noise_{\log}) + \log \mathrm{Pr}(\{\scaledx_{s}, \scaledy_{s}\}_{s=1}^{t}; \alpha_{\log}, \veclambdalog, \noise_{\log})\end{equation} 
for $ \alpha_{\log}, \veclambdalog, \noise_{\log}$ within the support of their respective prior distribution.

We use Scipy's L-BFGS-B to solve the constrained maximization problem, selecting the best results from four random initializations sampled from the hyperparameter priors. We sample the initial points from the uniform distribution of the truncated range.

\subsection{Acquisition Function and Trust Regions}
\label{subsec:acquisition}
We use a fixed UCB coefficient of $\sqrt{\beta} = 1.8$, which is relatively large compared to other open source settings \citep{hebo}. While the large coefficient can lead to better exploration, the downside is that regular GP-UCB would spend much of its initial budget exploring search space corners, as being furthest away from support points, they have the largest posterior variance estimates and thus highest acquisition values.

In order to control this behavior, we apply a trust region to numeric parameters, which is a union of $\ell_{\infty}$-balls around the ``trusted'' previously observed points $\{\scaledx_{s}\}_{s=1}^{t}$ in parameter space, or any points obtainable from those by changing categorical parameter values. 
The radius of these balls start at 0.2 in the scaled parameter space and grows unboundedly.
When optimizing the acquisition function, any evaluation outside the trust region gets strongly penalized, adjusted by the distance to the nearest trusted point. This is expressed as:
\begin{equation}
x \mapsto
\begin{cases}
\text{UCB}(x) & \text{if dist($\scaledx$, trusted) $\le$ radius}\\
-10^{12} - \text{dist($\scaledx$, trusted)} & \text{if dist($\scaledx$, trusted) $>$ radius}\\
\end{cases}
\end{equation}
The penalty value $-10^{12}$ is guaranteed to be lower than $\text{UCB}(x)$ due to our preprocessing steps and zero mean function. The additional distance term in the penalty is needed to provide ascent directions for acquistion optimizers to move into the trust region.

\subsection{Acquisition Function Optimization}
\label{subsection:acqusition_optimization}


We use a customized version of the Firefly algorithm \citep{firefly} where the mutation operators are customized per-datatype so that suggestions always correspond to points from the original $\X$.

Note that Ax \citep{botorch_official} by default uses L-BFGS-B for all-continuous spaces and a sequential greedy algorithm for mixed spaces, while HEBO \citep{hebo} uses NSGA-II \citep{nsga2}. Other alternatives include \textit{policy gradients} as in \citep{acqusition_gradient_softmax}.

\subsubsection{Firefly Algorithm}
The Firefly algorithm \citep{firefly} is a nature-inspired metaheuristic, particle swarm optimization technique \citep{pso} that draws inspiration from the flashing behavior of fireflies. In the original algorithm, a pool $\mathcal{P}$ of fireflies is maintained such that each firefly $\vecx$ represents a potential solution to the optimization problem, and its ``brightness" corresponds to the quality, or score $a(\vecx)$ of that solution. As the algorithm iterates, a ``dimmer firefly" $\vecx_{low}$ with a worse objective value will move towards towards a ``brighter'' $\vecx_{high}$, with the movement step's magnitude varying with distance $r(\vecx_{low}, \vecx_{high}) = \norm{\vecx_{low} - \vecx_{high}}_{2}$, which we shorthand as $r$. The individual firefly update rule is expressed formally in Equation \ref{eq:firefly_update}: 
\begin{equation} \vecx_{low} \leftarrow \vecx_{low} + \eta \exp(-\gamma \cdot r^2) \cdot (\vecx_{high} - \vecx_{low}) + \mathcal{N}(0, \omega^{2} I_{D}) \label{eq:firefly_update} \end{equation} 
where $\eta$ is the attraction coefficient, $\gamma$ is an an absorption coefficient and $\mathcal{N}(0, \omega^{2} I_{D})$ is an appropriately scaled random noise vector introduced to enhance exploration. The generic population update is expressed in Algorithm \ref{alg:generic_firefly}.

\vspace{0.3cm}
\begin{algorithm}[h]
\caption{Generic Firefly Population Update}\label{alg:generic_firefly}
\begin{algorithmic}
\For{$(\vecx_{i}, \vecx_{j}) \in \mathcal{P} \times \mathcal{P}$ s.t. $i < j$}
    \State Determine $(\vecx_{low}, \vecx_{high})$ by ranking $(x_{i}, x_{j})$ via $a(\cdot)$  
    \State $\vecx_{low} \gets \text{updated } \vecx_{low}$ using Eq. \ref{eq:firefly_update}
\EndFor
\end{algorithmic}
\end{algorithm}
\vspace{0.3cm}

In order to use the Firefly algorithm as an acquisition optimizer, we modified it with three crucial improvements: 

\begin{wrapfigure}{R}{0.5\textwidth}
    \centering
    \includegraphics[scale=0.46]{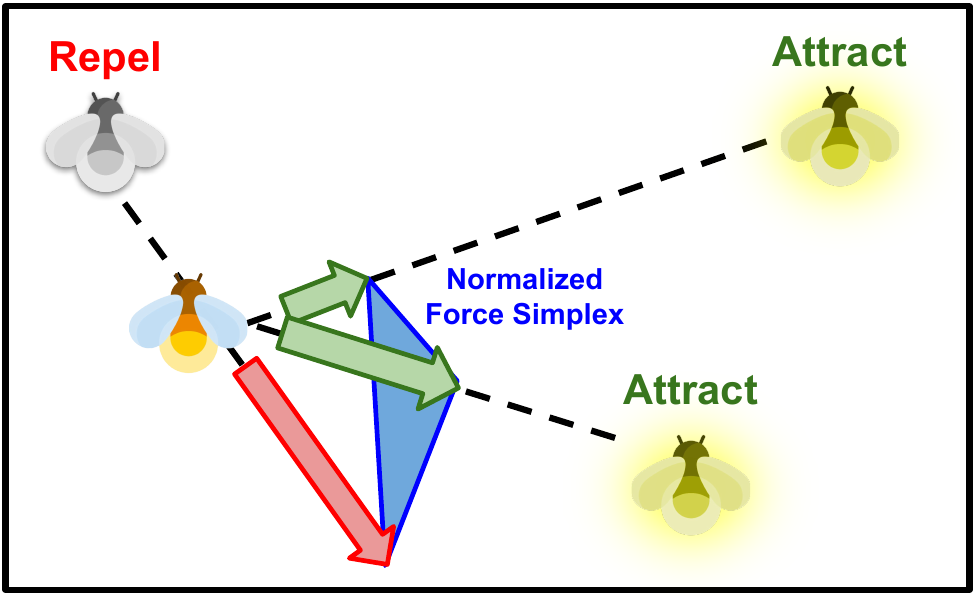}
    \caption{Visualization of vectorized firefly force normalization and summation. The resulting firefly will move into the simplex.}
    \label{fig:firefly_normalization}
\end{wrapfigure}

\textbf{Supporting Discontinuous Parameters:} To support discontinuous parameter types in Vizier, we round to one of the nearest values in $\X$ before evaluating the acquisition function. For \texttt{INTEGER} and \texttt{DISCRETE} parameters, we round to the nearest feasible value. For \texttt{CATEGORICAL} parameters, we interpret the mutated one-hot encoded vector as an unnormalized probability distribution from which we sample a single value.

\textbf{Utilizing Poor Mutations:} While the original Firefly algorithm emphasizes moving candidates toward superior solutions, 
we apply a repellent force from inferior fireflies on superior fireflies. 
The repellent forces are computed based on Equation \ref{eq:firefly_update}, where a negative value for the coefficient $\eta$ is used. Furthermore, fireflies trapped in poor local optima which consistently produce inferior evaluations are replaced with random candidates to promote greater search space exploration. 

\textbf{Vectorization and Batching:} Improving the wall-clock speed of the acquisition optimizer is crucial as it allows for more acqusition function evaluations and thus a better optimality gap. While the original C++ design multithreaded the Firely update, the new Python version vectorizes the algorithm to perform well on accelerators. Using JAX and just-in-time compilation, our vectorized method mutates a $\flybatchsize$-sized batch of fireflies by aggregating \textit{all} forces from the entire pool at once, and then applying normalization, i.e.
\begin{equation} \mathbf{\widetilde{X}} \leftarrow \mathbf{\widetilde{X}} + \frac{\eta}{\poolsize} \sum \left(e^{-\gamma \cdot r(\mathbf{\widetilde{X}}, \mathbf{X})^{2}} \odot (\mathbf{X} - \mathbf{\widetilde{X}}) \right) + \text{Laplace}(0, \omega) \label{eq:vectorized_firefly} \end{equation}
\begin{itemize}
\item $\mathbf{X} = \left[\vecx_{1}, \ldots, \vecx_{\poolsize}\right]^\top \in \R^{\poolsize \times D}$ contains the entire pool of fireflies, interpreting $\set{\vecx_i}$ as column vectors.
\item $\mathbf{\widetilde{X}} \in \R^{\flybatchsize \times D}$ is a batch of candidates to be updated in parallel.
\item $r(\mathbf{\widetilde{X}}, \mathbf{X})^{2} \in \R^{\flybatchsize \times \poolsize \times 1}$ contains pairwise squared distances from each firely in the batch to all fireflies in the entire pool.
\item $\mathbf{X} - \mathbf{\widetilde{X}} \in \R^{\flybatchsize \times \poolsize \times D}$ contains the pairwise feature increment of each firefly in the batch against all fireflies in the entire pool.
\item $\text{Laplace(mean, scale)}$ distribution is used for perturbations instead, due to better stability.
\item The multiplication $(\odot)$ is element-wise, the summation $(\sum)$ is on the dimension associated with the $\poolsize$ fireflies, and perturbations are sampled element-wise.
\end{itemize}
The normalization by the factor $\frac{1}{\poolsize}$ ensures that the accumulated fireflies force is contained within the simplex spanned by the individual firely forces. This prevents the fireflies from moving excessively which could potentially drive them outside the search space.

\subsection{Batched Setting}
\label{subsec:batch}
In many practical settings, users will request: (1) additional suggestions before all previous suggestions have finished being evaluated, or (2) a simultaneous batch of suggestions. In the context of Bayesian Optimization, algorithms for the first case can be used to incrementally construct batches for the second case. In either case, we must factor in previous suggestions as we wish to avoid accidentally (approximately) duplicating prior proposals. To do this, the algorithm looks at unevaluated suggestion features, and we define the GP via:
\begin{itemize}
\item \textbf{Mean function $\mu_{t}(\cdot)$}, which is still based on \textit{evaluated} trials $\mathcal{D}_{t} = \{x_{s}, y_{s}\}_{s=1}^{t}$, equivalent to the sequential case.
\item \textbf{Standard deviation function $\sigma_{t}(\cdot)$}, which \textbf{additionally} accounts for unevaluated suggestions $\mathcal{U}_{t} = \{x_{v},0\}_{v=1}$ given dummy zero objectives, an example of a ``constant liar'' heuristic \citep{constant_liar}. 
\end{itemize}
Following the ideas of \cite{contal2013parallel} but modified to improve performance and align with infrastructure, we augment our GP-UCB algorithm with \textit{pure exploration}, where we define two acquisition functions:
\begin{itemize}
\item \textbf{UCB (Upper Confidence Bound):} 
\begin{equation} \label{eqn:ucb-def}
\mbox{UCB}(x | \mathcal{D}_t, \mathcal{U}_t, \beta) := \mu_t(x|\mathcal{D}_t) + \sqrt{\beta} \cdot \sigma_t(x|\mathcal{D}_t \cup \mathcal{U}_t)
\end{equation}
\item \textbf{PE (Pure Exploration):}
\begin{equation} \label{eqn:pe-def}
\mbox{PE}(x | \mathcal{D}_t, \mathcal{U}_t, \tau_t, \beta_e, \rho) := \sigma_t(x|\mathcal{D}_t \cup \mathcal{U}_t) + \rho \cdot \min\left(\mbox{UCB}(x | \mathcal{D}_t, \emptyset, \beta_e) - \tau_t, 0\right)
\end{equation}
where $\rho$ is a penalty coefficient, $\emptyset$ denotes empty set, $\beta_e$ is an exploration-specific UCB parameter, and $\tau_t$ is a threshold on the UCB value for exploration, computed solely conditioned on evaluated trials $\mathcal{D}_{t}$.
\end{itemize}
The PE acquisition function encourages controlled exploration by maximizing predicted standard deviation $\sigma_{t}$ within a promising region where the predicted UCB values are reasonably high. The threshold $\tau_t$ that defines the promising region is computed as $\tau_t := \mu_t(x_t^*|\mathcal{D}_t)$, where $x_t^*$ is the suggestion with the highest standard UCB value, i.e.
\begin{equation}
x_t^* = \argmax_{x \in \mathcal{D}_t \cup \mathcal{U}_t} \> \> \mbox{UCB}(x | \mathcal{D}_t, \emptyset, \beta)
\end{equation}
To generate a batch of suggestions, the algorithm dynamically selects between the two acquisition functions to guide its exploration. Specifically, the algorithm optimizes (1) UCB when generating a suggestion following new trial evaluations, and (2) PE when generating suggestions without new trial evaluations. However, with small probability during case (1), we still use PE instead to ensure exploration. 

Algorithm \ref{alg:gp_ucb_pe} in the Appendix concisely demonstrates the process of generating a batch of suggestions. The trust region from Section \ref{subsec:acquisition} around previously evaluated points may optionally be applied, restricting the acquisition optimization domain for enhanced stability.

\subsection{Multi-objective optimization}
\label{subsec:multiobjective}
In multiobjective optimization, $\Y = \mathbb{R}^{M}$, and $f$ is a vectorized function of scalar functions, i.e. $f(x) = (f^{(1)}(x), ..., f^{(M)}(x))$. For $y_1, y_2 \in \Y$, we say $y_2$ \emph{Pareto dominates} $y_1$, written as $y_1 \prec y_2$ if for all coordinates $i$, $y_1^{(i)} \le y_2^{(i)}$ and there exists $j$ with $y_1^{(j)} < y_2^{(j)}$. For two points $x_1, x_2 \in \X$, we say that $x_2$ {\it Pareto-dominates} $x_1$ if $f(x_1) \prec f(x_2)$.
A point is {\it Pareto-optimal} in $\X$ if no point in $\X$ dominates it. Let $\X^* \subset \X$ denote the set of Pareto-optimal points in $\X$, and correspondingly $\Y^* = f(\X^\star)$ denote the {\it Pareto frontier}.

The goal of multiobjective optimization is to find the Pareto frontier. Our main measure of progress is the hypervolume that is Pareto dominated by the metric vectors obtained so far.
Specifically, for a compact set $S \subseteq \R^M$, let $\operatorname{vol}(S)$ be the hypervolume of $S$. For a set of metric vectors $Y \subset \R^M$, we use the (dominated) {\bf hypervolume indicator} of $Y$ with respect to a \emph{reference point} $y_{\text{ref}}$ as our progress metric:
\begin{equation} \HV_{y_{\text{ref}}}(Y) = \operatorname{vol}\paren{\set{ u \,  |  \,  
y_{\text{ref}} \le u \text{ and } \exists y \in Y \text{ such that }  u \prec y}}
\end{equation}
A natural approach for maximizing the hypervolume is to greedily maximize the hypervolume gain at every step $t$, leading to widespread use of the Expected Hypervolume Improvement (EHVI) acquisition and its differentiable counterpart \citep{daulton2020differentiable}. Unfortunately, computation of the hypervolume indicator is \textbf{\#P}-hard in general~\citep{hypervolume_nphard}. 

\subsubsection{Hypervolume Approximation via Scalarization} 
Practitioners often solve multiobjective problems by choosing \emph{scalarization functions} from metric vectors to scalars, and optimizing the latter. 
Let $s_w: \mathbb{R}^M \to \R$ denote a family of \emph{scalarization functions} parameterized by $w$.
By optimizing $s_w(f(x))$ for various $w$ to obtain the next suggestion, we effectively reduce the multiobjective case to the single objective case.
A key observation is that when $s_w$ is monotonically increasing with respect to all coordinates in $y$, $\arg\max_{x} \set{s_w(f(x))}$ is on the Pareto frontier. 

We utilize the particular family of 
{\bf hypervolume scalarizations} to ensure the converse: each point on the Pareto frontier equals $\arg\max_{x} \set{s_w(f(x))}$ for some $w$. 
Additionally, the Pareto frontier can be provably estimated in an unbiased manner by sampling $w$ and averaging the optimizers of the hypervolume scalarizations:

\begin{theorem}[\cite{golovin2020random}]
\label{lem:hypervolume}
Let $\set{y_1,..., y_k}$ be a set of points in $\R^M$. Then, the hypervolume of this set with respect to a reference point $y_{\textup{ref}}$ is given by:
\begin{equation} \HV_{y_{\text{ref}}}(\set{y_1,..., y_k}) = c_M \cdot \mathbb{E}_{w} \left [ \max_{y\in \set{y_1,..., y_k}} s_w(y - y_{\textup{ref}} ) \right ] \label{eq:golovin2020} \end{equation}
where $s_w(y) = \Big(\underset{1 \le m \le M}{\min} \textup{ReLU}(y^{(m)}/w^{(m)}) \Big)^M$, $w$ is drawn uniformly from $\set{u \, | \, u \in \R^M_{> 0}, ||u||_2 = 1}$,
and $c_M = \frac{\pi^{M/2}}{2^M \ \Gamma(M/2+1)} $ depends only on $M$.
\end{theorem}

\begin{figure}[t]
\begin{center}
\centerline{\includegraphics[height=2.2in]{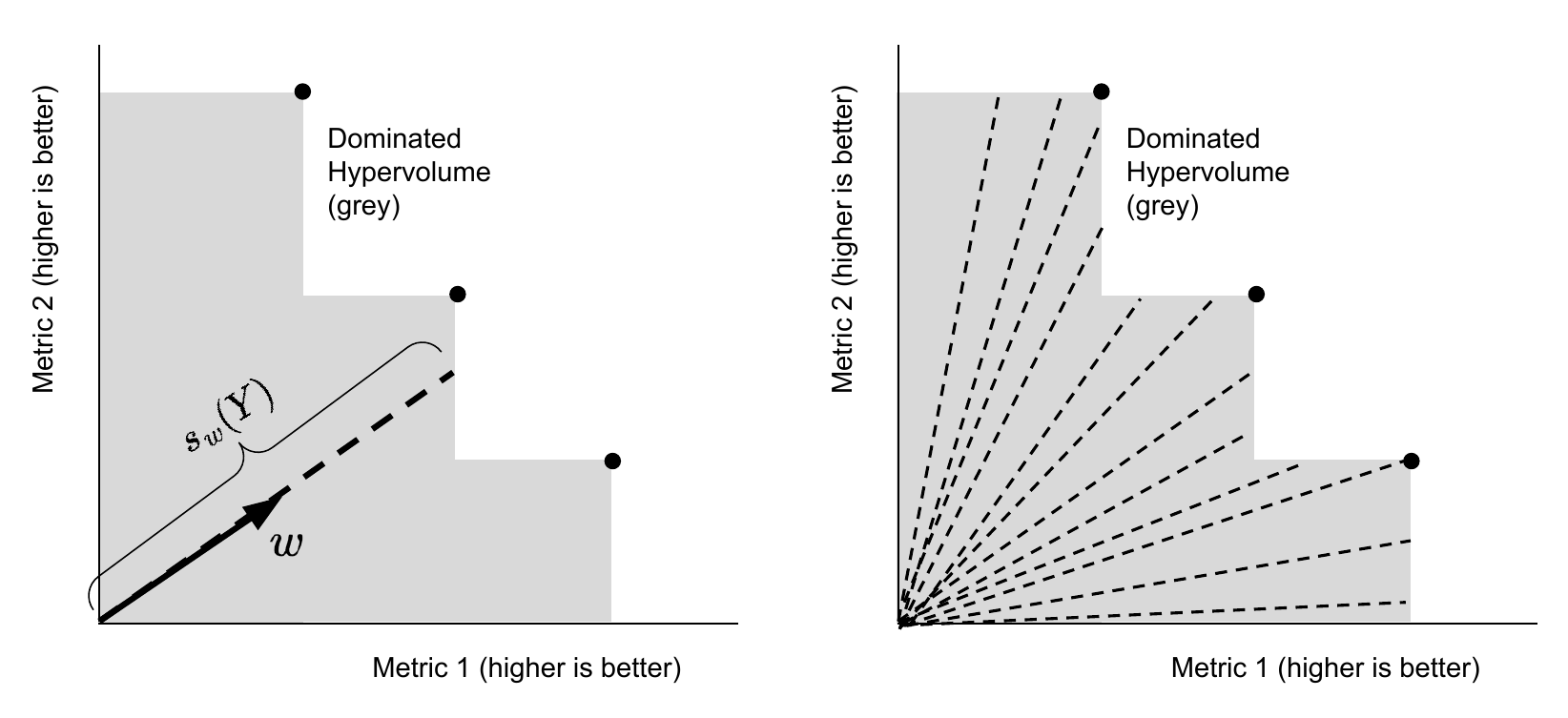}}
\caption{\small \textbf{Left:} The hypervolume scalarization taken with respect to a direction $w$ corresponds to a differential area  (dashed line) within the hypervolume dominated by set of metric vectors $Y$ (black dots). Here, $s_w(Y) := \max \set{s_w(y) : y \in Y}$. \textbf{Right:} Averaging over random directions is analogous to integrating over the dominated hypervolume in polar coordinates (up to a dimension-dependent constant).\vspace{-1.2cm}}
\label{fig:hv_polar}
\end{center}
\end{figure}

\subsubsection{Acquisition Scalarization and Multi-task GP} 
Based on our approximation method above, a natural way to produce an acquisition for multiobjective posteriors is to apply scalarizations on a vectorized per-metric upper confidence bound  $\overrightarrow{\operatorname{UCB}}(x) = (\operatorname{UCB}^{(1)}(x), \ldots, \operatorname{UCB}^{(M)}(x))$. Then, we proceed to perform classic maximization on the average scalarized acquisition $\mathbb{E}_{w} \left[s_w(\overrightarrow{\operatorname{UCB}}(x))\right]$ with hypervolume scalarizations, which is related to the hypervolume of the upper confidence vector via Theorem~\ref{lem:hypervolume}, and can be approximated by averaging over thousands of random draws of weights $w$, due to fast JAX linear algebraic operations. 

Since we are measuring the \textit{improvement} of the cumulative hypervolume upon adding a point $x$, our acquisition must substract the maximized scalarized value $\max_{y \in \mathcal{D}_t} \set{s_w(y)}$ across previous trials in the dataset $\mathcal{D}_{t}$. We find that using hypervolume-scalarized vector UCB works well in practice, as compared to other choices of acquisition such as EHVI, and believe that its bias towards optimism helps find better exploration and exploitation tradeoffs in hypervolume maximization. Furthermore, this method is provably optimal for minimizing hypervolume regret \citep{zhang2023optimal}:
\begin{equation}
x_{t+1} = \argmax_{x \in \X} \> \> \mathbb{E}_{w} \left[\max\left\{0, s_w(\overrightarrow{\operatorname{UCB}}(x)) - \max_{y \in \mathcal{D}_t} \set{s_w(y)} \right\} \right] \label{eq:hypervolume_scalarized_vector_UCB}
\end{equation}
Note the reference point $y_{\text{ref}}$ is an implicit part of the acquisition function. We determine it using lower bounds on each metric individually (as detailed in Appendix~\ref{sec:exact-hps}). 

To model each of the $y^{(i)}$, we use the same GP modeling assumptions as in the single-objective case, with the only difference being that this multi-task GP allows for modeling correlations between objectives, which can lead to improved gains in partially observable or noisy data regimes and for well-constructed objectives. For sake of simplicity and practicality, by default we use the \textit{Independent kernel}, which simply performs a Kronecker product between the base Matern kernel and the identity matrix. Note that our acquisition choice is informed by our choice of correlation modeling.

\subsection{Trial Seeding and Initialization}
\label{subsec:seeding_and_initialization}
\begin{wrapfigure}{R}{0.5\textwidth}
    \centering
    \includegraphics[scale=0.16]{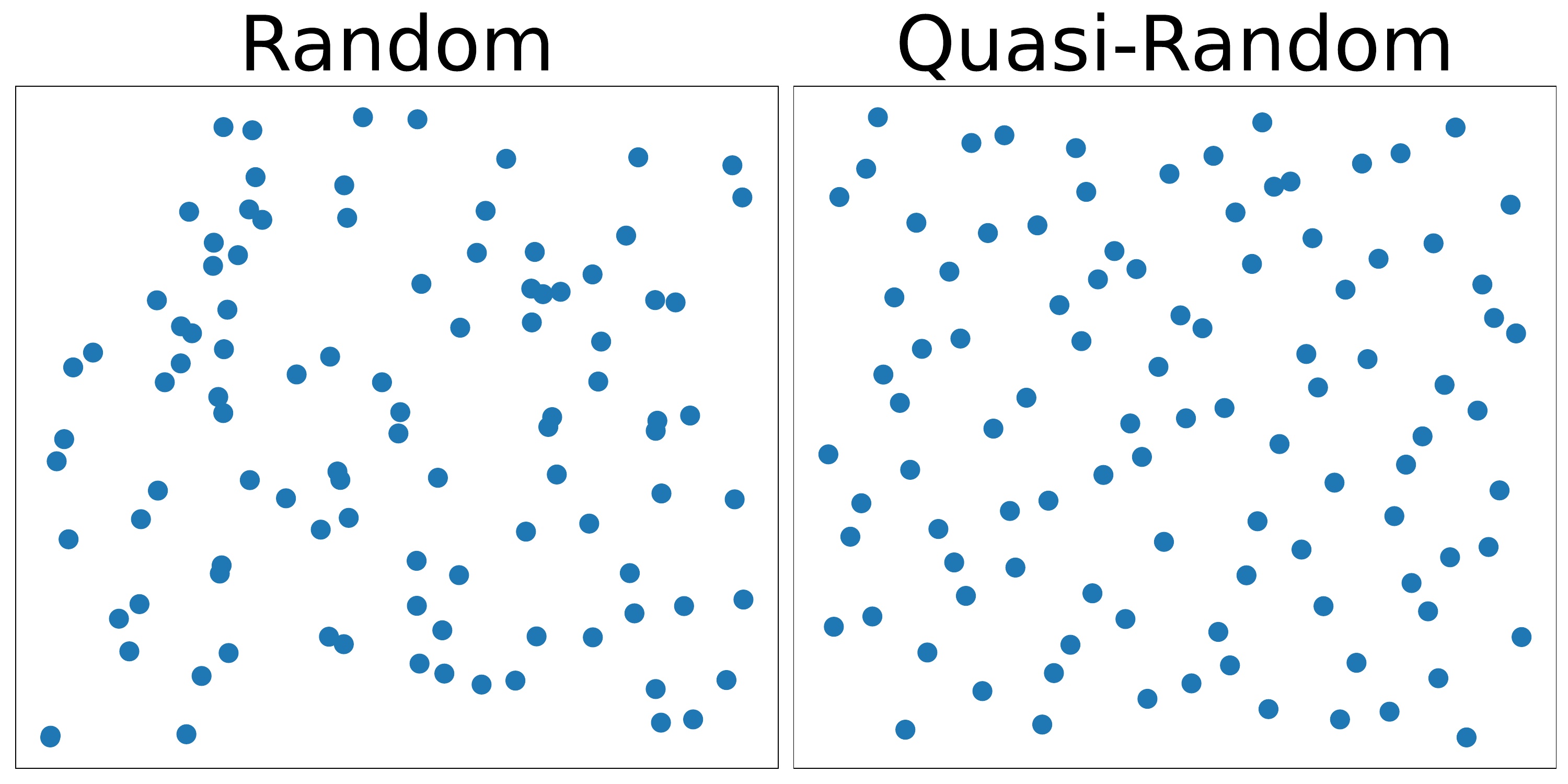}
     \caption{Example of i.i.d. random sampling vs. quasi-random sampling over a 2D space.}
\end{wrapfigure}

In order for the Gaussian Process to obtain a rich spread of evaluations over the search space, we use a different selection for the first few trials. 

\textbf{Initial Centering:} In many practical cases, users tend to define search spaces which contain the optimum. We fix the initial trial to be the center of the search space (uniform sampling for \texttt{CATEGORICAL} parameters) which obtains a relatively close guess to this optimum, especially for high dimensions.

\textbf{Quasi-Random Search:} Afterwards, we may optionally sample trials quasi-randomly using Halton sequences \citep{halton} as opposed to i.i.d. randomly, in order to obtain a more uniform spread over the search space, with the number of quasi-random trials proportional to the parameter count of $\mathcal{X}$.

\section{Experiments: Ray Tune Comparisons}
We compare the Vizier algorithm with well-established baselines published on Ray Tune \citep{raytune}: Ax/BoTorch \citep{botorch_official}, BayesianOptimization \citep{bayesian_optimization_github}, HEBO \citep{hebo}, HyperOpt \citep{hyepropt}, Optuna \citep{optuna}, and Scikit-Optimize \citep{scikit_optimize}. These baselines are all variants of Bayesian optimization, but mostly differ in their acquisition functions and optimizers. A summary of these differences are in Appendix \ref{appendix:baseline_details}.

\subsection{Evaluation Protocol}
Our emphasis for this paper is on production-quality and user accessibility, implying a stronger focus on robustness and out-of-the-box behavior without the need for knob-tuning. Thus we present results using default settings for all algorithm baselines in our main body, with Appendix \ref{appendix:more_experiments} containing comparisons when certain knobs are changed. Using defaults also aligns with the intentions of algorithm developers, who may provide their own algorithm selectors (e.g., Ax selects its algorithm based on search space). We further diversify our collection of benchmark functions (Appendix \ref{appendix:benchmark_objectives}) to avoid possible issues if certain algorithms were tuned for any specific benchmarks. 

\textbf{Trajectory Generation:} Unless otherwise specified, every optimization trajectory is run with 100 trials with 20 repeats, and we plot the median curve along with 40-60 percentile error bars among all figures.



\textbf{Log-Efficiency Violin Plots:} To get algorithm comparisons aggregated over multiple objective functions, we use the \textit{log-efficiency} metric, explained further in Appendix \ref{appendix:evaluation_protocol}. In summmary, for a Ray Tune algorithm $\mathcal{A}$, we define its $\text{RequiredBudget}(y \> | \> f, \mathcal{A} )$ with respect to an objective $f$ and a target value $y$ as the minimum number of iterations required to reach or surpass $y$. Then we may define \textit{relative log-efficiency} with respect to Vizier, as:
\begin{equation}
    \text{LogEfficiency}(y \> | \> f, \mathcal{A}) = \log\left(\frac{\text{RequiredBudget}(y \> | \>  f, \text{Vizier})}{\text{RequiredBudget}(y \> | \> f, \mathcal{A} )} \right)
\end{equation}
Our final log-efficiency score for a pair $(\mathcal{A}, f)$ is the median of LogEfficiency's when ranging $y$ across the averaged best-so-far trajectories between $\mathcal{A}$ and Vizier. Each baseline's final scores across different objectives $f_{1}, f_{2}, \ldots$ are plotted as distributions in violin plots (with 25-75 quartiles), with Vizier represented by a zero horizontal axis.

\textbf{Initial Trial Centering:} We empirically found that proposing the search space center produces a drastic initial gap in performance across many benchmarks. To make comparisons fairer, we always inject the centering trial at the beginning of a trajectory, with the exception of random search to demonstrate the gap.

\subsection{Purely Continuous Spaces}
We begin by benchmarking over continuous spaces in a sequential manner where trials are suggested and evaluated one-by-one. This setting is the most researched and supported by all baselines.

\vspace{0.2cm}
\begin{figure}[h]
    \centering
    \includegraphics[width=0.95\textwidth]{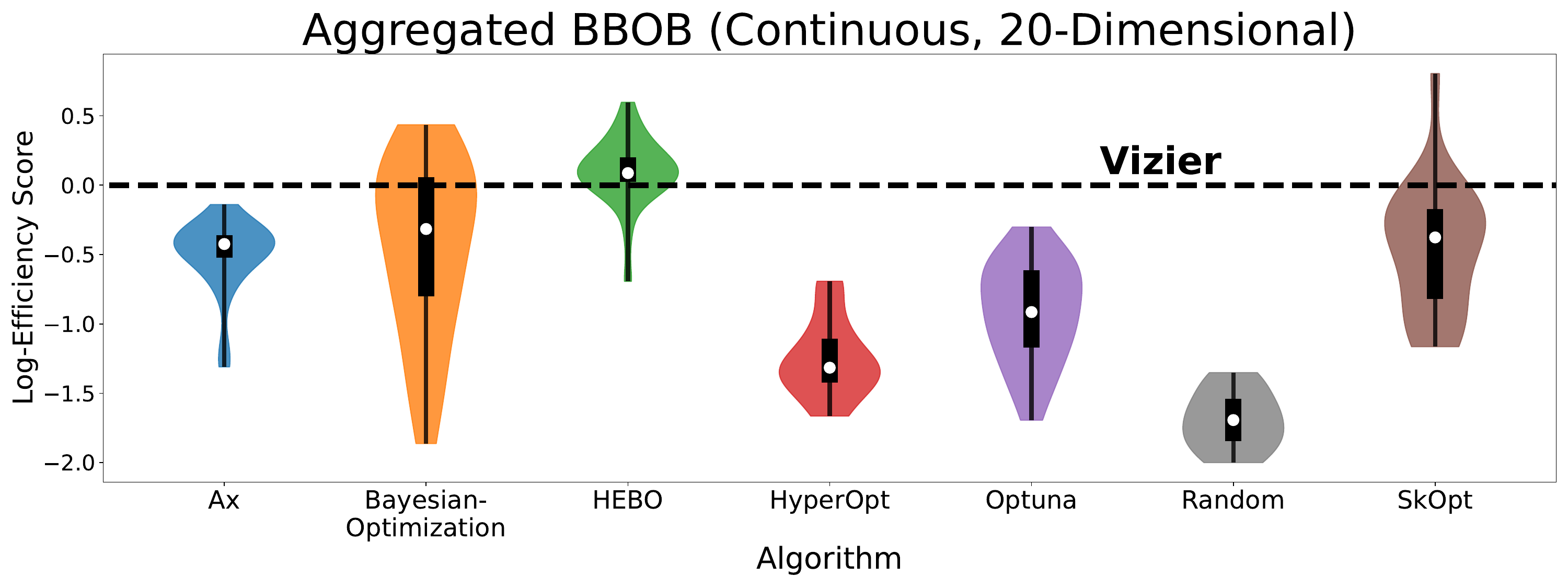}
    \caption{\small Higher is better $(\uparrow)$. Violin plots displaying distributions of log-efficiency scores across all BBOB functions. Similar plot for noisy objectives in Appendix \ref{appendix:noisy_bbob}.\vspace{0.3cm}}
    \label{fig:continuous_20d_violin}
\end{figure}

\begin{wrapfigure}[15]{r}{0.5\textwidth}
\vspace{-0.3cm}
    \centering
    \includegraphics[width=0.48\textwidth]{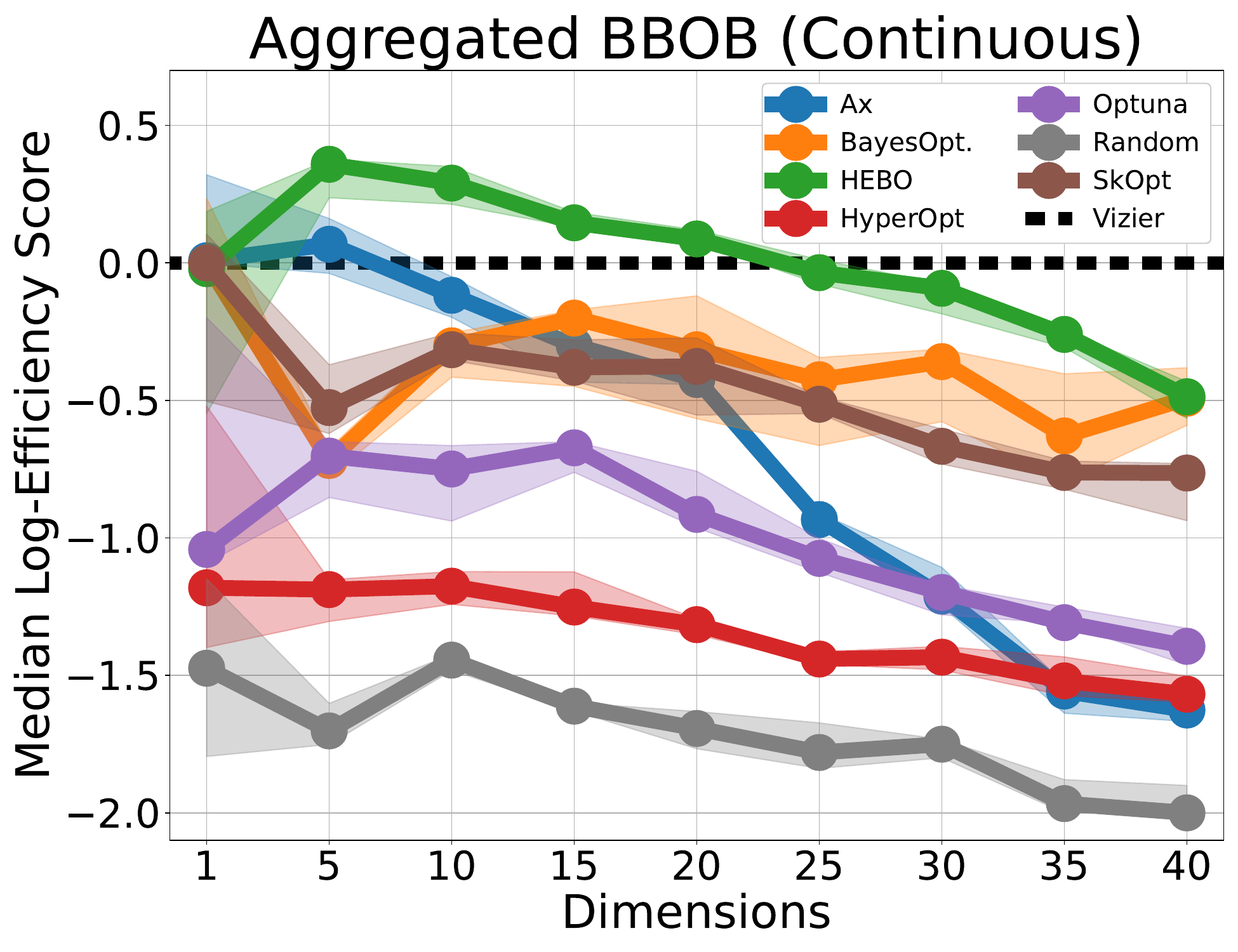}
    \caption{\small Higher is better $(\uparrow)$. Median log-efficiency scores across algorithms over continuous BBOB functions, while varying search space dimensionality.}
    \label{fig:continuous_varying_dim}
\end{wrapfigure}

In order to comprehensively assess general performance on various objective landscapes and scales, we benchmark over various transformations of the standard Blackbox Optimization Benchmark (BBOB) suite \citep{bbob}. We set the search space dimension to be relatively high (e.g. 20), as we found lower-dimensional problems insufficient for distinguishing the benefits of Vizier's components (e.g. warping, MAP estimation, and acquisition optimization).

\begin{figure}[t]
    \centering
    \includegraphics[width=\textwidth]{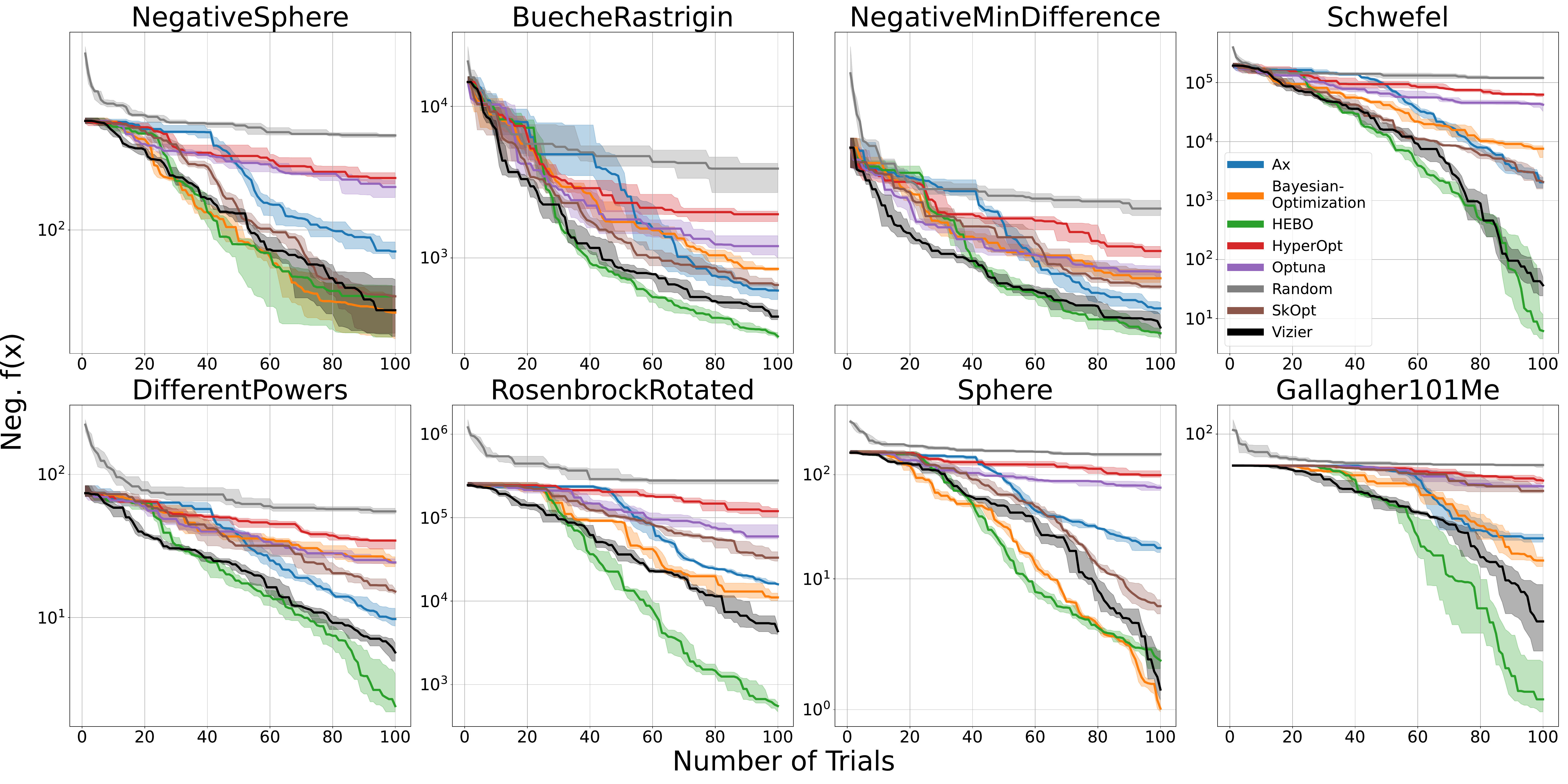}
    \caption{\small Lower is better $(\downarrow)$. Optimality gap curves across 8 randomly chosen 20-dimensional BBOB functions. Note: y-axis is log-scaled to depict clearer separation between baselines.}
    \label{fig:continuous_20d_8_plots}
\end{figure}

We first provide an overall distribution of log-efficiency scores via a violin plot in Figure \ref{fig:continuous_20d_violin} and demonstrate Vizier's relatively competitive performance. We also provide a fine-grained view of individual BBOB results by plotting best-so-far curves over varying functions in Figure \ref{fig:continuous_20d_8_plots}. We see that for later trials, HEBO is a strong competitor in many cases, followed by BayesianOptimization and Ax. In Appendix \ref{appendix:more_experiments}, we demonstrate that Ax's median log-efficiency remains the same even when using the same acqusition as Vizier's UCB ($\sqrt{\beta}$ = 1.8).

However, in Figure \ref{fig:continuous_varying_dim} we find that these baselines begin to underperform, sometimes drastically (e.g. Ax), as we increase the problem dimensionality. While studies usually have smaller search spaces, especially for users familiar with selecting important hyperparameters to tune, many real world use cases can be high dimensional. Such cases occur frequently when users are unaware of said hyperparameters in advance, and thus create a high dimensional search space by adding all possible parameters for Vizier to optimize.

\subsection{Non-Continuous Spaces} 
Many important hyperparameters and tuning decisions are inherently non-continuous (e.g. choosing to use neural networks vs. random forests), and therefore assessing performance over categorical spaces is crucial.

\vspace{0.3cm}
\begin{figure}[h]
    \centering
    \includegraphics[width=\textwidth]{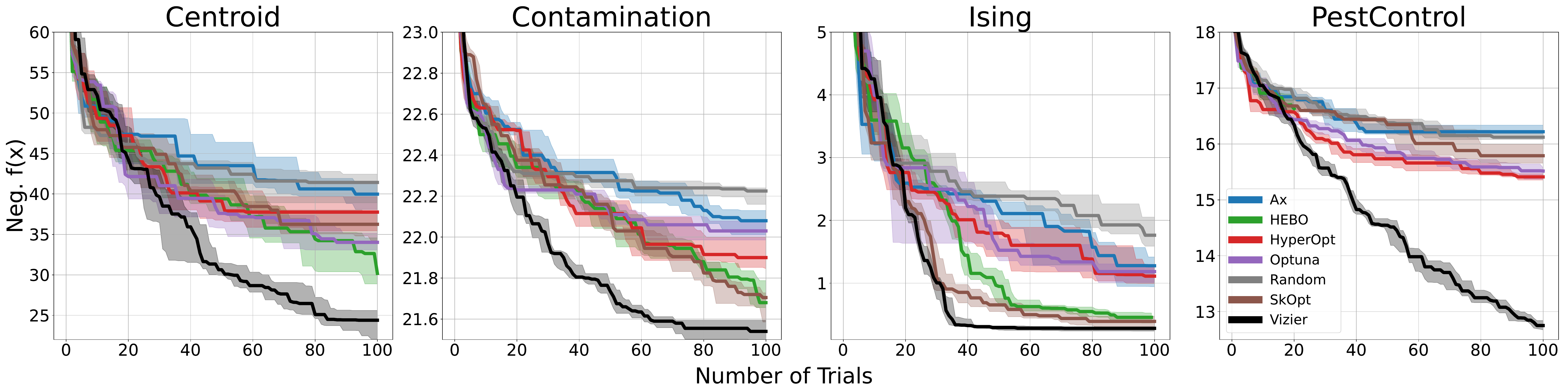}
    \caption{Lower is better $(\downarrow)$. Best-so-far across categorical objectives (\textbf{Centroid:} 24D $\times$ 3 Categories, \textbf{Contamination:} 25 Booleans, \textbf{Ising:} 20 Booleans, \textbf{PestControl:} 25D $\times$ 5 Categories).  Note: HEBO halted early in PestControl.\vspace{0.3cm}}
    \label{fig:combo_4_plots}
\end{figure}

Using operations research objectives introduced in \cite{combo}, we assess the capabilities of algorithms over ordinal and categorical search spaces in Figure \ref{fig:combo_4_plots}. We find that Vizier consistently outperforms other methods over high dimensional categorical spaces, especially for later trials. 

\begin{wrapfigure}{R}{0.5\textwidth}
    \centering
    \includegraphics[width=0.48\textwidth]{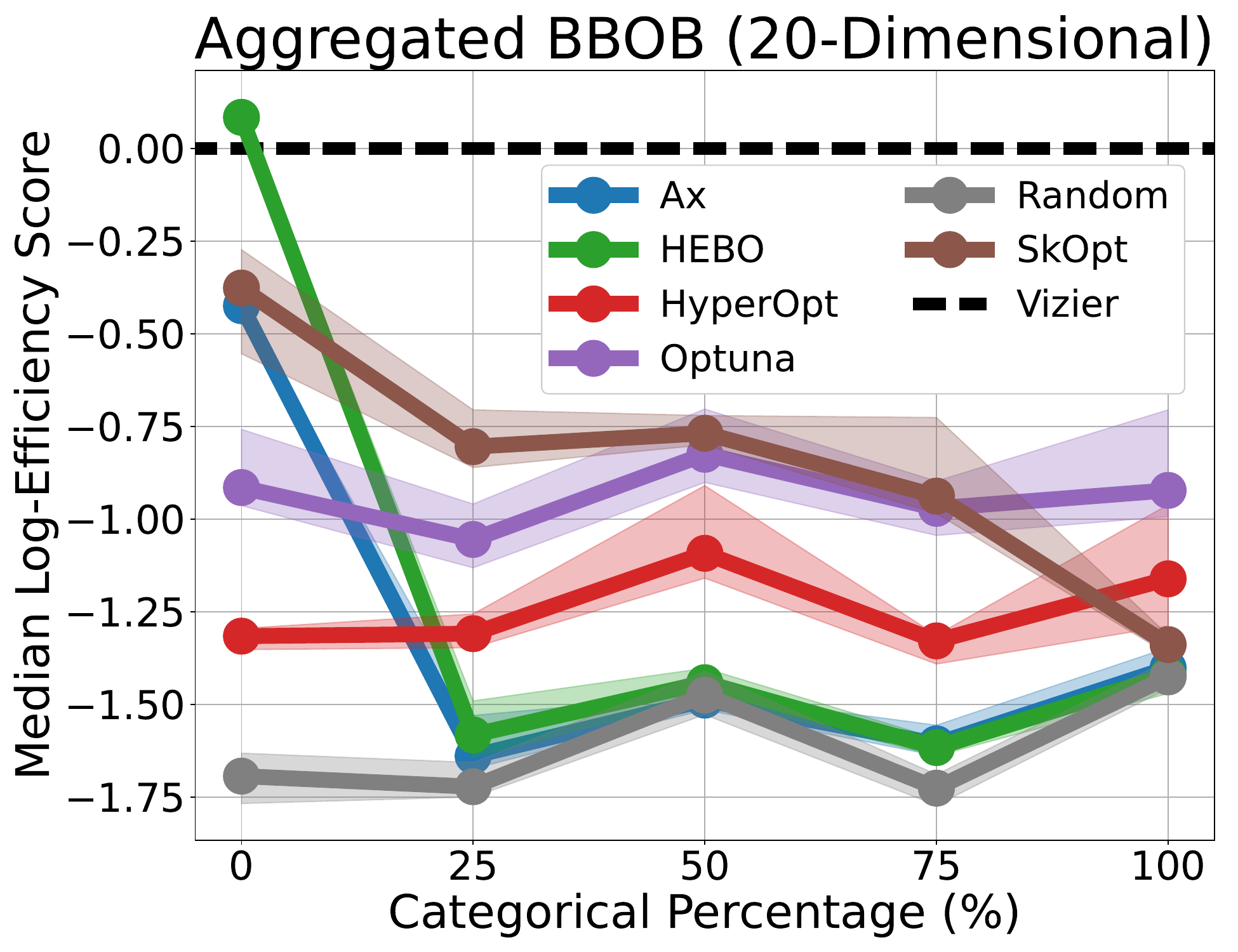}
    \caption{\small Higher is better $(\uparrow)$. Median log-efficiency scores across categorical-supported algorithms over BBOB functions, while varying number of categorical parameters in the search space.}
    \label{fig:varying_hybrid}
\end{wrapfigure}

Furthermore, in many use-cases, the search space may contain a hybrid mix of continuous and categorical variables. As a convenient benchmarking method, we can ``categorize'' any of the continuous parameters originally from our BBOB objectives by selecting 10 equidistant grid points as feasible values. In Figure \ref{fig:varying_hybrid}, we demonstrate Vizier significantly outperforms other baselines even when a few (25\%) categorical parameters exist. We consistently found that for search spaces containing high category parameters, HEBO halts early due to Cholesky decomposition errors in its Gaussian process, demonstrating categorical modeling can be challenging, in addition to the acquisition optimization issues mentioned in Section \ref{subsection:acqusition_optimization}.

\subsection{Batched Case}
\begin{figure}[h]
    \centering
    \includegraphics[width=0.8\textwidth]{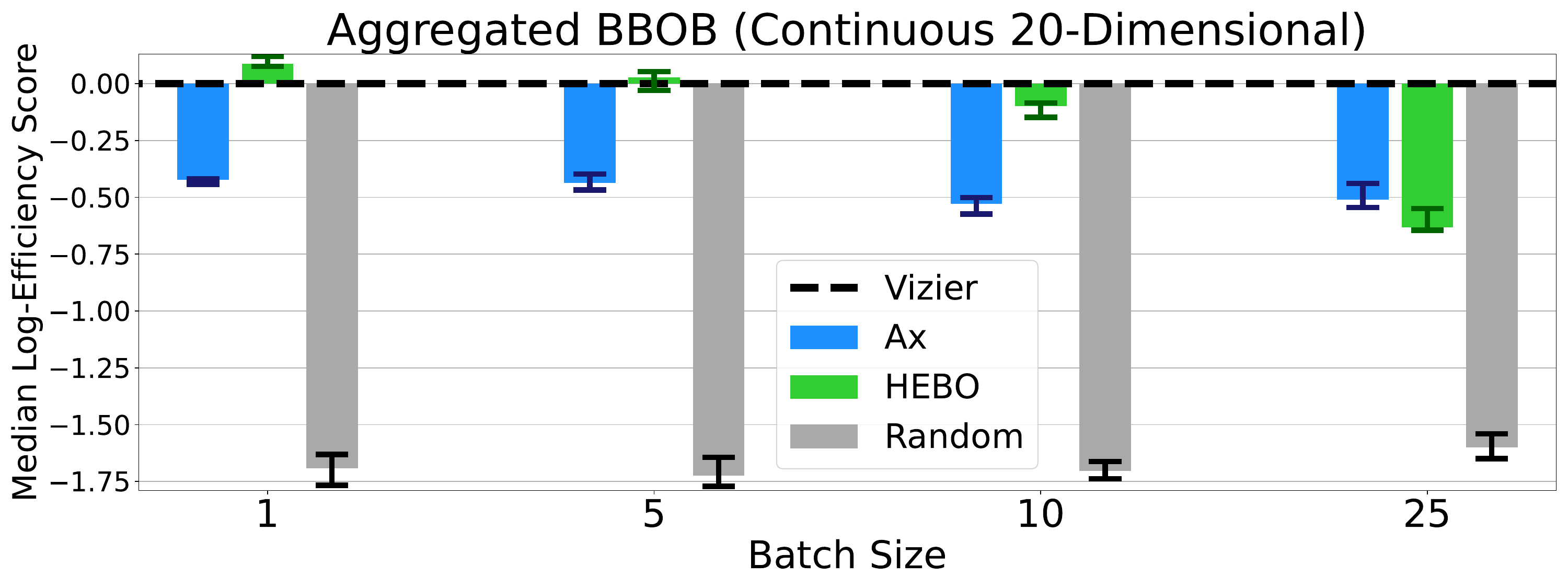}
    \caption{Higher is better $(\uparrow)$. Median log-efficiency scores across algorithms over BBOB functions while varying batch size.}
    \label{fig:varying_batch}
\end{figure}

Given the above sequential results, it suffices to benchmark the best performing baselines with batched capabilities, namely Ax and HEBO. We vary the batch size (i.e. \texttt{``max\_concurrent\_trials''} in the Ray Tune API) and see that in Figure \ref{fig:varying_batch}, even for large batches, Vizier still remains efficient against random search, while HEBO's performance in particular degrades, demonstrating the importance of our pure exploration mechanism from Section \ref{subsec:batch}.

\vspace{0.3cm}

\subsubsection{Multi-Objective Case}

We directly use Ax, HEBO, and Optuna via their official API for benchmarking comparisons, as multi-objective optimization is not directly supported in Ray Tune. Our problems consist of collections of well-known synthetic functions designed specifically for multi-objective optimization, specifically ``DTLZ'' \citep{dtlz_multiobjective}, ``WFG'' \citep{wfg_multiobjective}, and ``ZDT'' \citep{zdt_multiobjective}, totalling 21 different functions. All functions allow varying dimensions, and with the exception of ZDT, also the number of objectives. Since previous experiments already establish Vizier's strength in high dimensional optimization, we instead focus on multi-objective capabilities, by fixing the dimension of $\X$ to be relatively small.

\begin{figure}[t]
    \centering
    \includegraphics[width=0.95\textwidth]{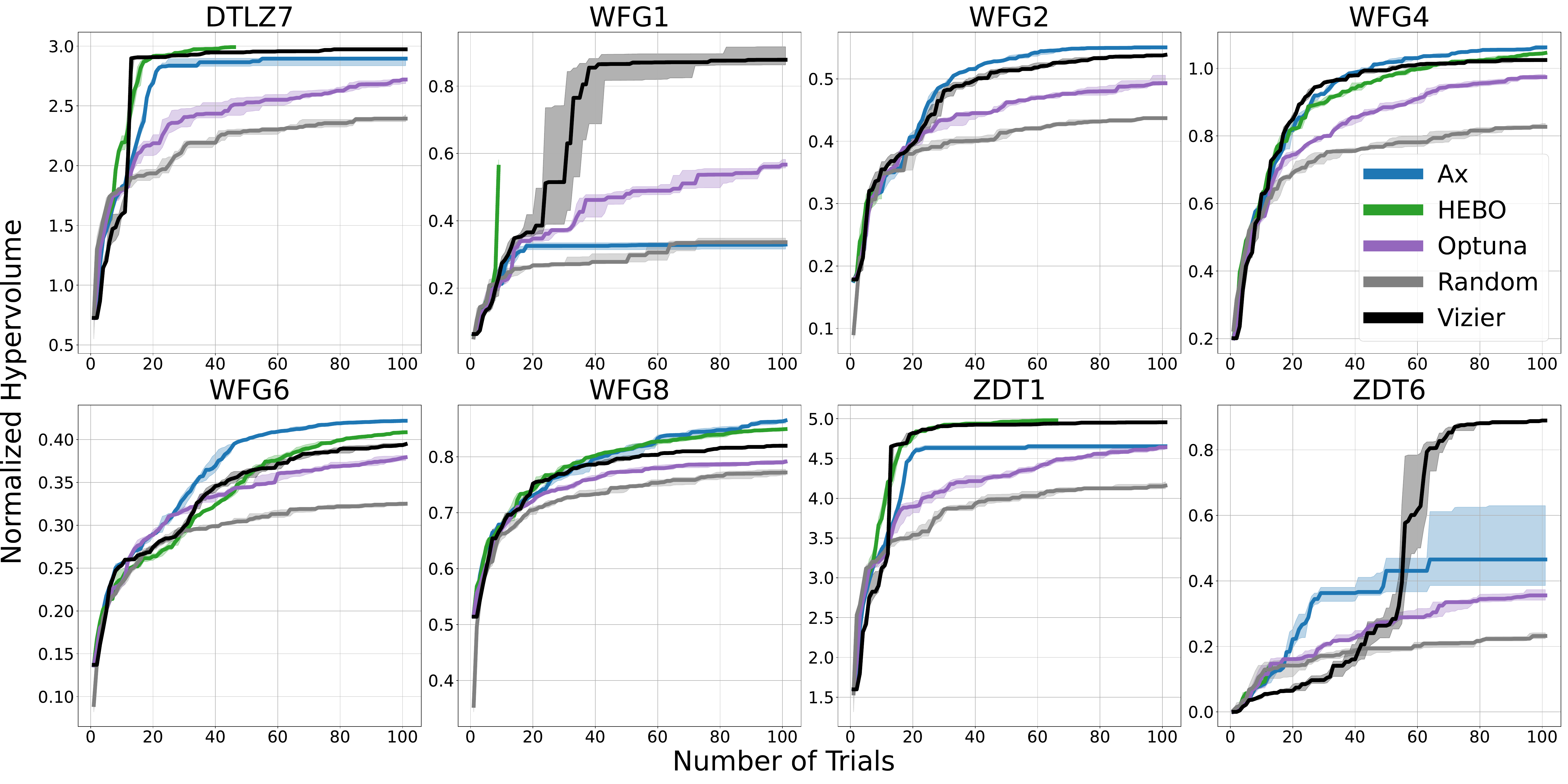}
    \caption{Higher is better $(\uparrow)$. Randomly selected individual normalized hypervolume plots over 5-dimensional functions (DTLZ, WFG, ZDT) with 2 objectives.\vspace{-0.4cm}}
    \label{fig:individual_multiobjective}
\end{figure}

\begin{figure}[h]
  \centering
  \begin{minipage}[b]{0.46\textwidth}
    \includegraphics[width=\textwidth]{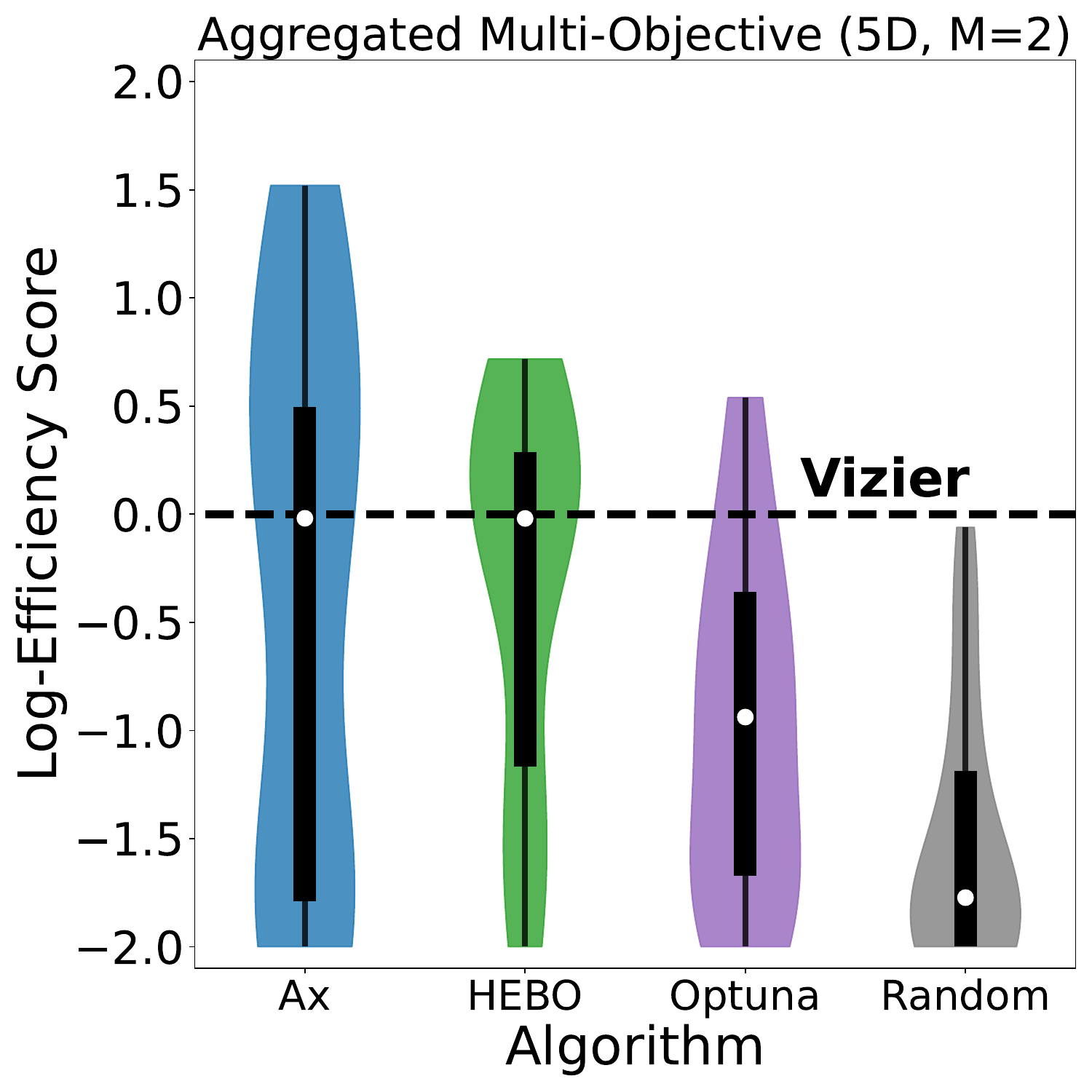}
  \end{minipage}
  \begin{minipage}[b]{0.46\textwidth}
    \includegraphics[width=\textwidth]{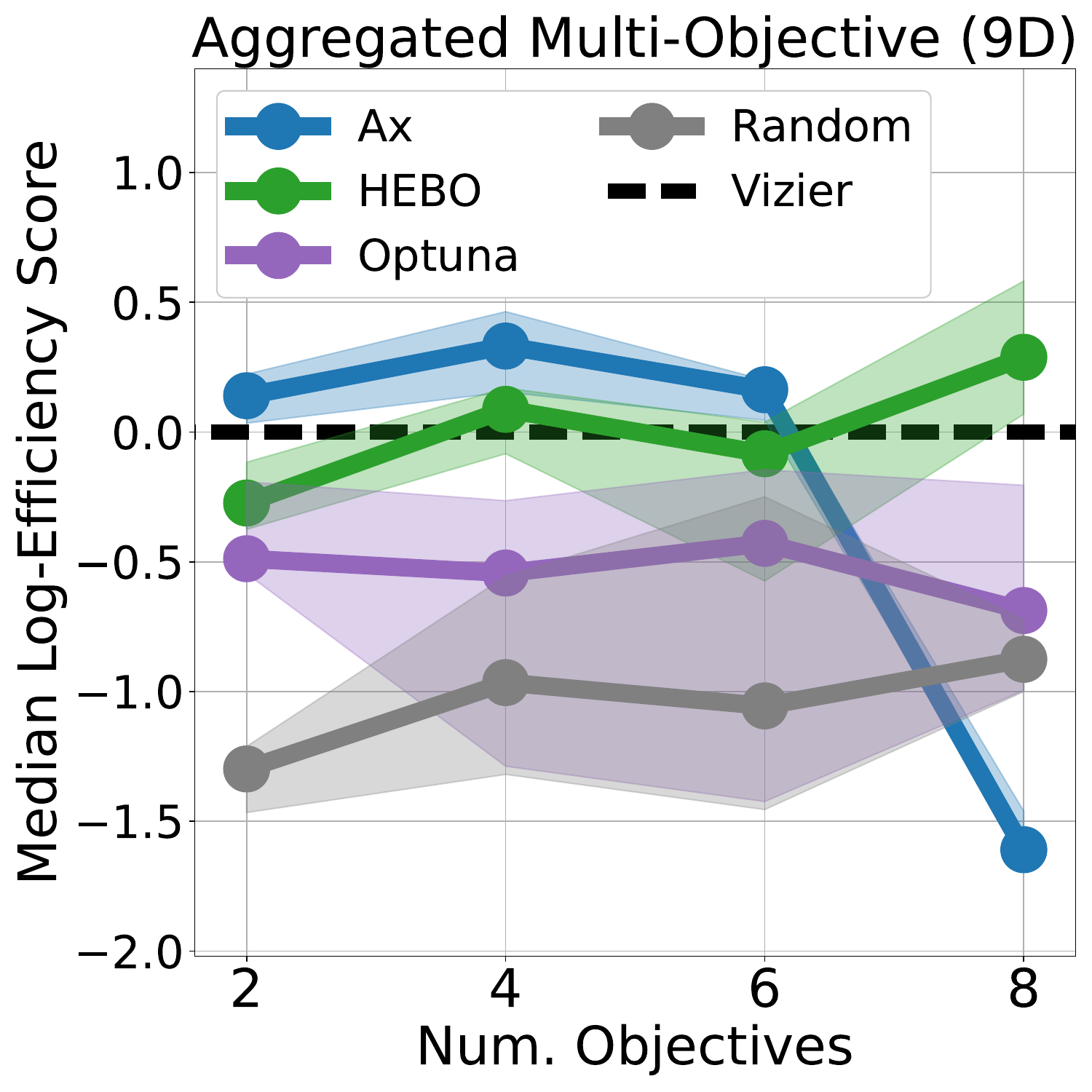}
  \end{minipage}
    \caption{\small \textbf{Left:} Higher is better $(\uparrow)$. Violin plots of log-efficiencies across algorithms over multi-objective functions (DTLZ, WFG, ZDT). \textbf{Right:} Higher is better $(\uparrow)$. Median log-efficiency scores across algorithms over multi-objective functions (DTLZ, WFG), while varying number of objectives.\vspace{0.2cm}}
    \label{fig:multiobjectives}
\end{figure}

In Figure \ref{fig:multiobjectives} (Left), we see that when comparing over the aggregate median, Vizier performs similarly against Ax and HEBO, while significantly outperforming Optuna and random search. However, we also find a few outlier cases in which the rankings of algorithms can wildly fluctuate. Some evidence of this can be seen individually in Figure \ref{fig:individual_multiobjective}, where e.g. Ax underperforms in WFG1 and Vizier initially underperforms in ZDT6. We hypothesize these fluctuations are due to usage of different acquisition functions. We further see again, numerical instability issues halting HEBO early in e.g. DTLZ7, WFG2, and ZDT6.

\vspace{0.1cm}

In Figure \ref{fig:multiobjectives} (Right), we vary the number of objectives $M$ as an additional stress-test, simulating another ``kitchen-sink'' scenario in which some users define numerous metrics to understand the trade-offs between all of them. We see that while HEBO remains stable throughout, Ax suffers substantially over high objective counts, again demonstrating Vizier's competitive robustness.

\section{Experiments: Ablations}
\subsection{Firefly vs. L-BFGS-B for Acquisition Optimization}
The acquisition function landscape can have many local maxima, which under some conditions can be comparable to the number of support points. Vectorized Firefly handles such a landscape well, by simultaneously exploring multiple promising regions of the search space efficiently over O(10K) points. In contrast, second-order algorithms such as L-BFGS-B (details in Appendix \ref{appendix:baselines}) make overly strong assumptions about acquisition function landscape shape and plateau too early.

\vspace{0.2cm}
\begin{figure}[h]
    \centering
    \includegraphics[width=0.8\textwidth]{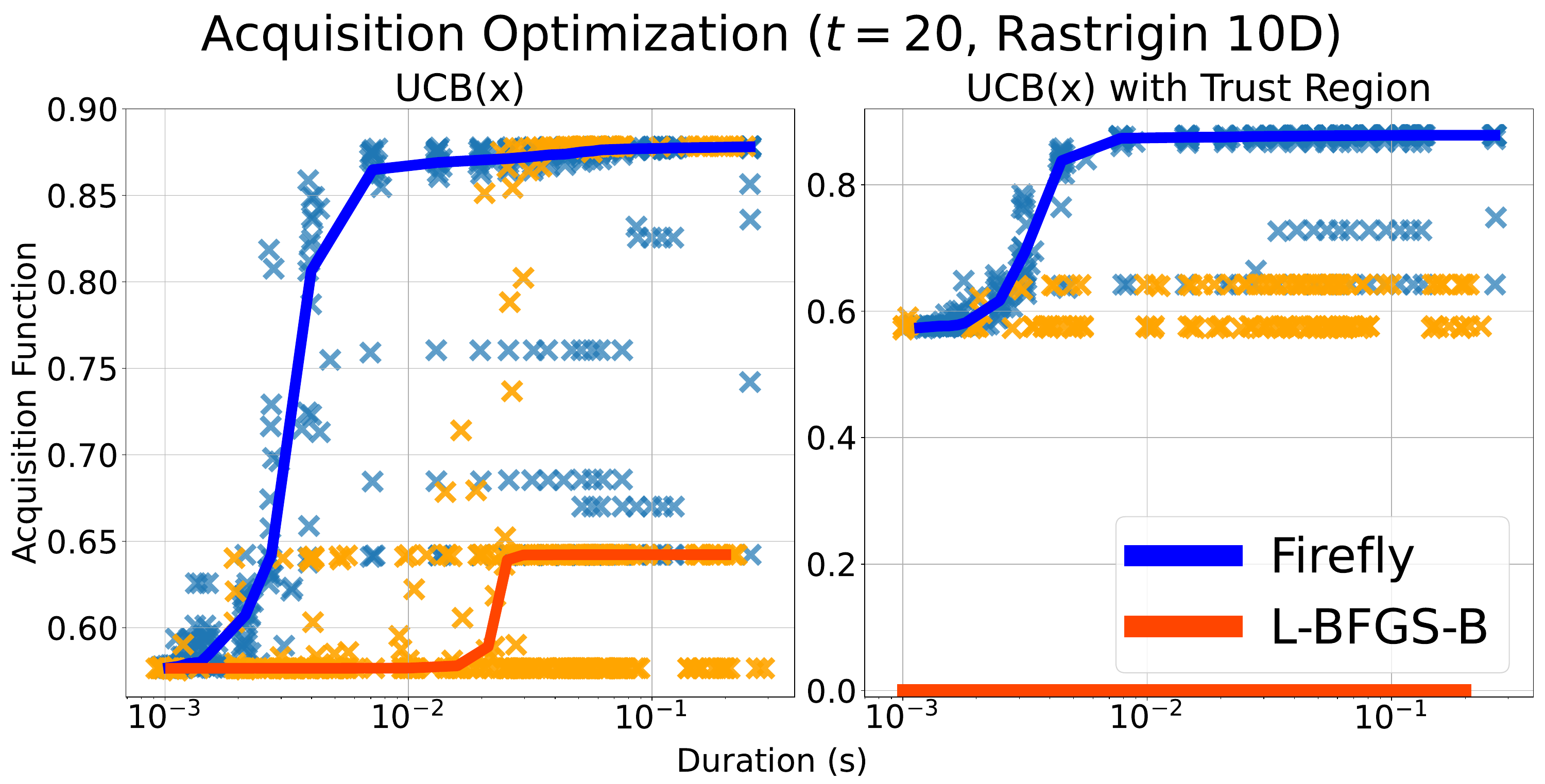}
    \caption{\small Upper-left is better $(\nwarrow)$. Firefly vs. L-BFGS-B on UCB(x), from a randomly chosen benchmark's optimization loop. Both optimizers ran with varying iteration counts to generate scattered points (light blue \& orange), alongside their median curves (dark blue \& orange). \textbf{Note:} L-BFGS-B's median curve actually obtained the trust region penalty of $-10^{12}$ on the right, but is clipped to 0 for easier visualization.
    \vspace{0.3cm}
    }
    \label{fig:eagle_vs_lbfgsb_acquisition}
\end{figure}
 
In Figure \ref{fig:eagle_vs_lbfgsb_acquisition}, we directly demonstrate that Firefly consistently optimizes the acquisition function better at every wall-clock budget, while L-BFGS-B plateaus early on. An observation is that when trust regions are used, L-BFGS-B significantly struggles to enter the trust region, while Firefly has no such trouble. 

\vspace{0.3cm}
\begin{figure}[h]
    \centering
    \includegraphics[width=1.0\textwidth]{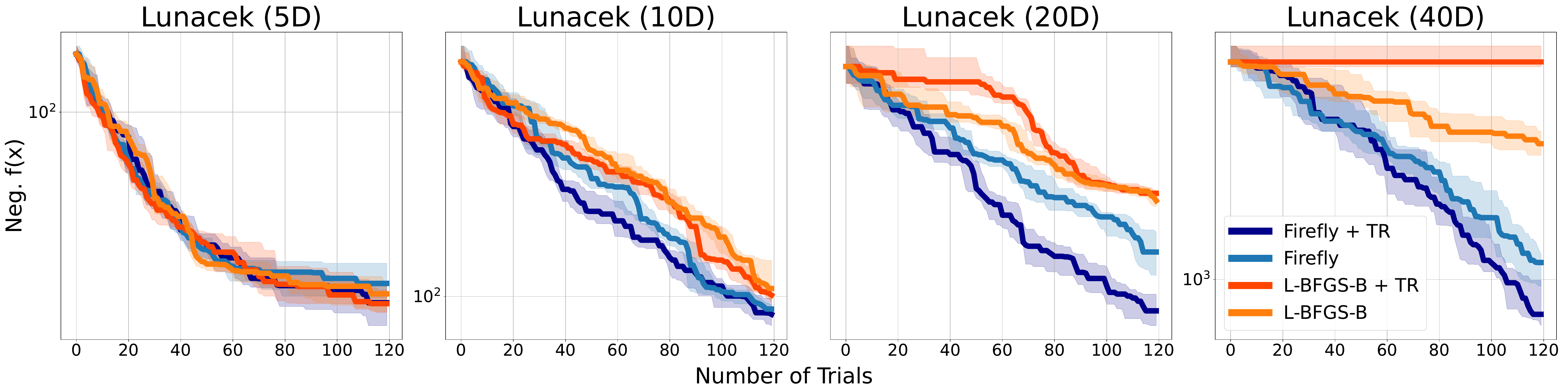}
    \caption{Lower is better $(\downarrow)$. Firefly vs. L-BFGS-B, with and without trust region (TR) usage on end-to-end Bayesian optimization process on the (randomly shifted) Lunacek BBOB function.\vspace{0.3cm}}
    \label{fig:e2e_bayesopt_eagle_vs_lbfgsb}
\end{figure}

As seen in Figure \ref{fig:e2e_bayesopt_eagle_vs_lbfgsb}, Firefly substantially leads to better performance at higher dimensions, but is boosted even further by the use of trust regions, made possible by the versatility of Firefly in successfully optimizing different acquisition function implementations. These results demonstrate that the Firefly algorithm is well-suited for Vizier's user-facing applications, which prioritize both rapid convergence towards optimal solutions, low suggestion latency, and robustness over different parameter types.


\subsection{Latency and GPU Acceleration}

\begin{wrapfigure}[18]{r}{0.5\textwidth}
    \centering
    \includegraphics[width=0.45\textwidth]{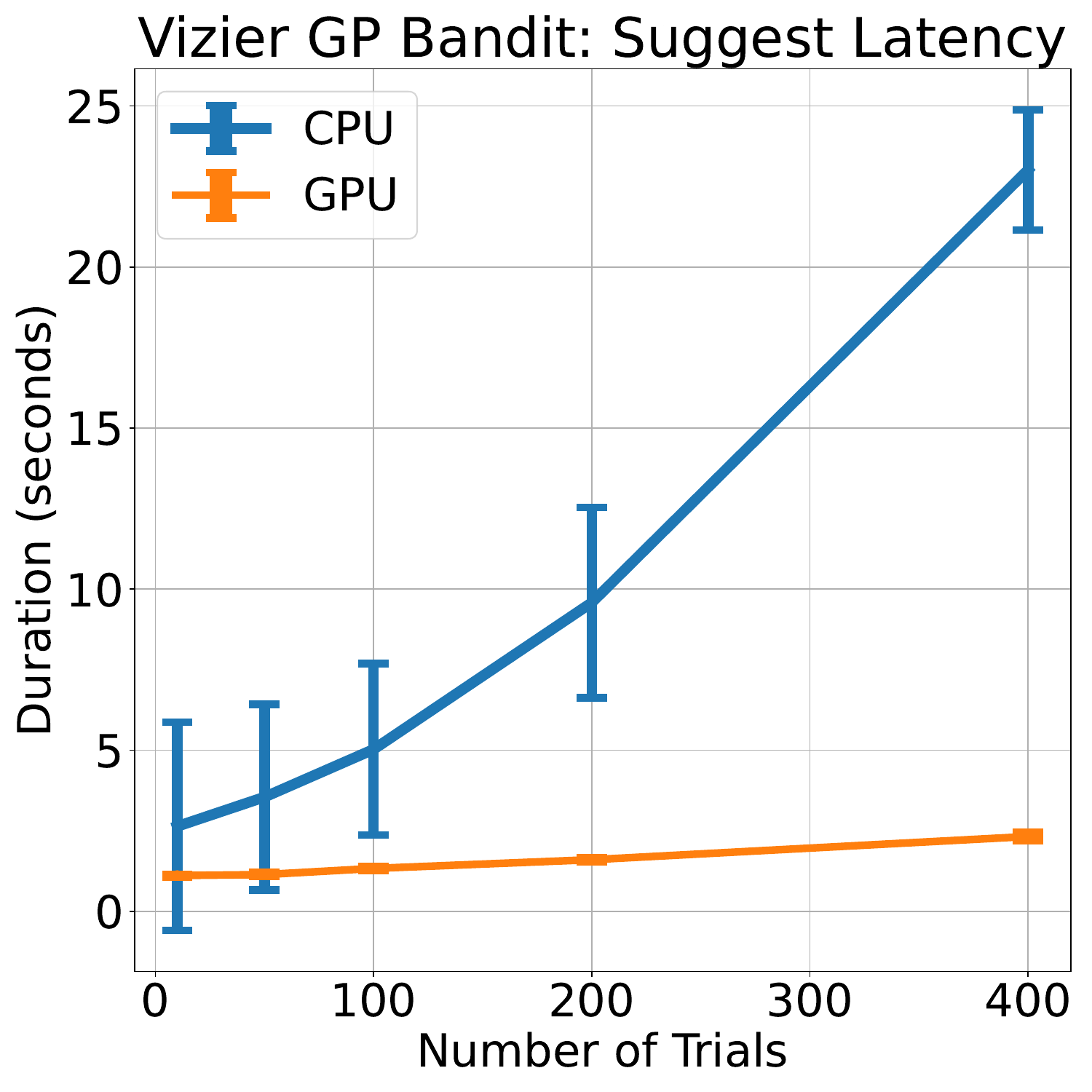}
    \caption{Lower is better $(\downarrow)$. Vizier GP-Bandit suggestion latency on CPU vs. GPU, measured as a function of history length. Experiment was conducted with an 8-D Rastrigin function.}
    \label{fig:cpu_vs_gpu}
\end{wrapfigure}

Since components in the GP-bandit algorithm were implemented in JAX and optimized with JIT-compilation, wall-clock speedups can be obtained when using accelerators. In Figure \ref{fig:cpu_vs_gpu}, we see that the algorithm runs significantly faster on GPU compared to CPU, making serving the algorithm quite cheap.

\section{Conclusion}

We have provided the Vizier default algorithm implementation details, which can be organized in terms of trial preprocessing, response surface modeling with Gaussian process prior and kernel, acquisition function definition, and evolutionary acquisition optimization. Throughout our experiments, we have demonstrated the algorithm's robustness over other industry-wide baselines across multiple axes, namely non-continuous parameters, high-dimensional spaces, batched settings, multi-metric objectives, and even numerical stability. This robustness has proven to support a wide variety of use-cases within Google and we expect it to additionally be an invaluable tool for the external research community.



\section*{Acknowledgements}
We thank Jacob Burnim and Brian Patton for help in writing TF Probability code and solving bugs. We further thank Zi Wang for providing feedback on this paper, Claire Cui, Zoubin Ghahramani, and Jeff Dean for continuing support.



\clearpage
\bibliography{refs}

\clearpage
\appendix

\section{Extended Experiments}
\label{appendix:more_experiments}
\subsection{Comparison with Ax's UCB}
\begin{wrapfigure}{R}{0.5\textwidth}
    \centering
    \includegraphics[width=0.45\textwidth]{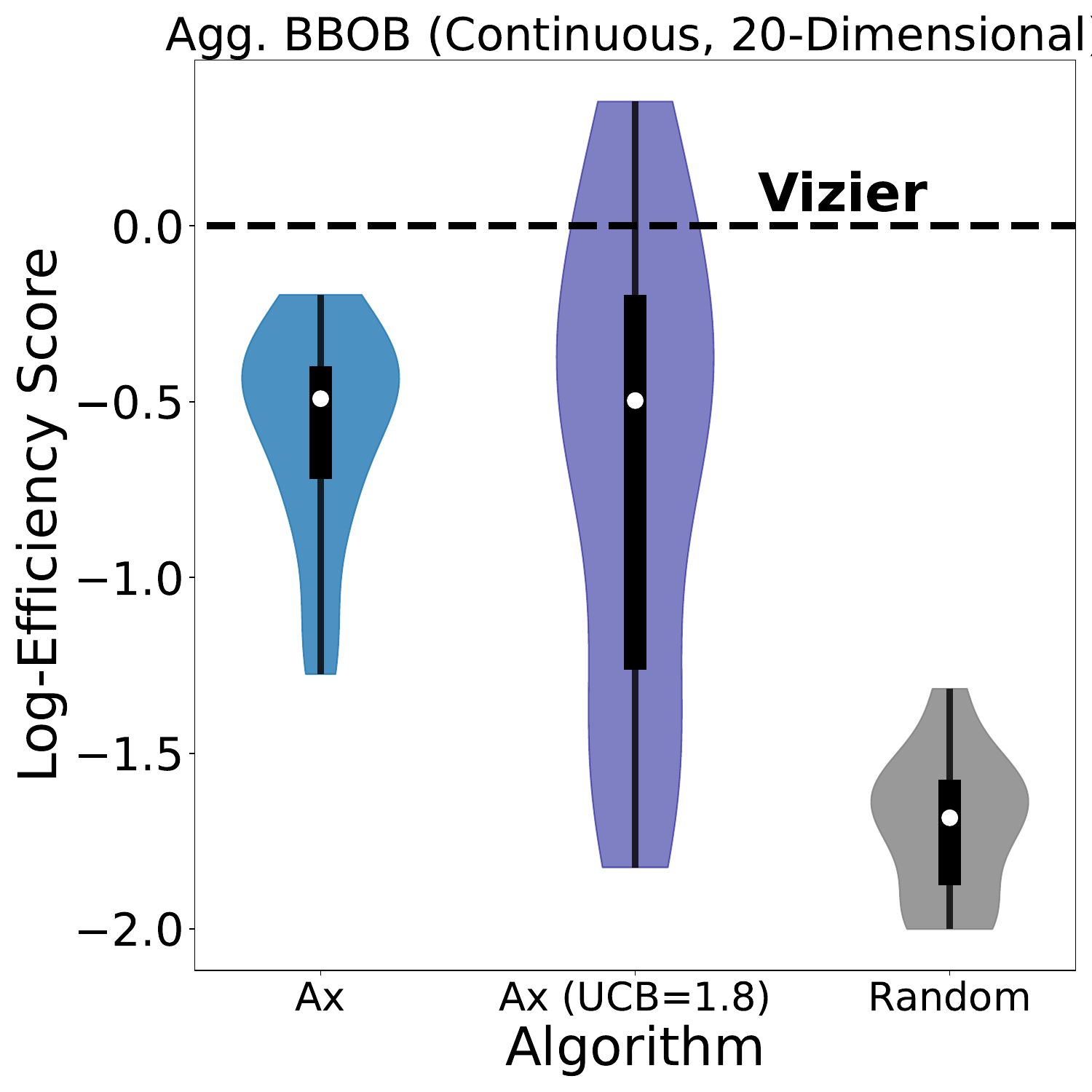}
    \caption{Higher is better $(\uparrow)$. Violin plots displaying distributions of log-efficiency scores across all BBOB functions.}
    \label{fig:ax_ucb_violin}
\end{wrapfigure}
One question raised by the main experiments is whether the performance gaps between Vizier and e.g. Ax are purely due to differences in acquisition function definition, as Ax uses the family of expected improvement (EI) acquisitions \citep{bayesian_ei, ax_ei} by default, rather than UCB. In order to resolve this hypothesis, we further modify Ax to use the \texttt{UpperConfidenceBound} acquisition method with $\sqrt{\beta}$=1.8, similar to Vizier's UCB. We further disable this variant's \texttt{SOBOL} (i.e. quasi-random) sampling for an even more direct comparison to Vizier's single-objective method.

\begin{figure}[h]
    \centering
    \includegraphics[width=1.0\textwidth]{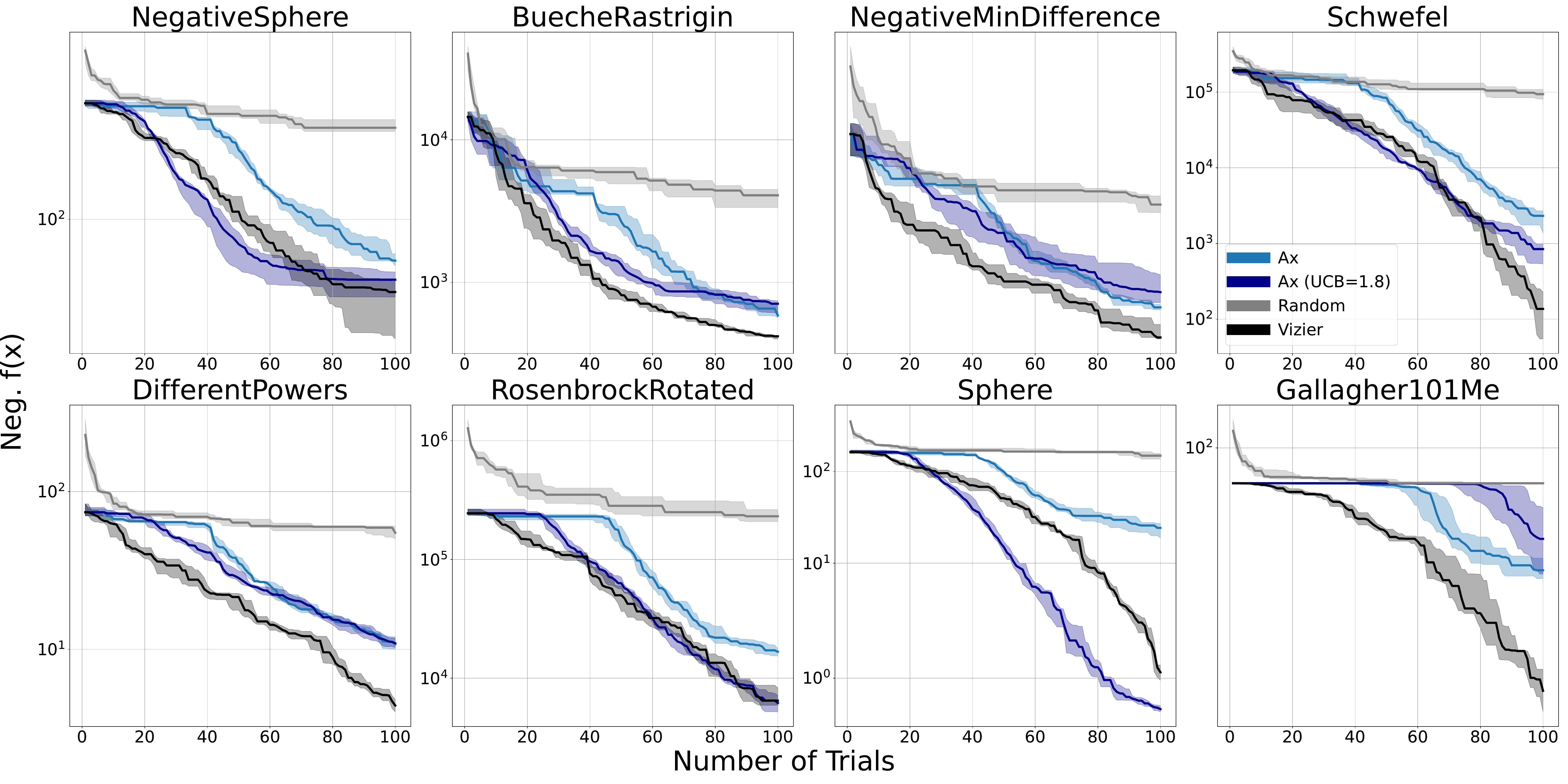}
    \caption{Lower is better $(\downarrow)$. Best-so-far curves across 8 randomly chosen 20-dimensional BBOB functions. Note: y-axis is log-scaled to depict clearer separation between baselines.}
    \label{fig:ax_ucb_individual}
\end{figure}

In Figure \ref{fig:ax_ucb_violin}, we see that the median roughly remains the same for Ax-based methods, regardless of acquisition function definition, although UCB leads to a higher variance in performance. In Figure \ref{fig:ax_ucb_individual}, we see some of these individual cases, and underperformance is most prounounced in some functions such as Gallagher101Me where both variants of Ax are unable to improve in the first half of the trial budget.

These results highlight the importance of other components within Vizier (e.g. acquisition maximization, kernel definition, and prior hyperparameter optimization) which also contribute to its robustness and success.

\vspace{0.2cm}

\subsection{Noisy Objectives}
\label{appendix:noisy_bbob}
Many real world evaluations tend to be noisy, i.e., repeated evaluations of $f(x)$ at a fixed $x$ may lead to different values. A typical mathematical abstraction is to assume that $f(x)$ inherently produces a distribution of values over $\mathbb{R}$ from which measurements $y$ may be sampled.

For BBOB functions, there is a standard set of ``noise models'' which can be wrapped over the original deterministic functions, as prescribed in \cite{noisy_bbob}. These consist of two multiplicative (Gaussian and Uniform) and an additive (Cauchy) noise model, which are described in Appendix \ref{appendix:benchmark_objectives}. While the algorithm observes the noisy value per evaluation, our definition of performance however is still based on the original deterministic value.

\begin{figure}[h]
    \centering
    \includegraphics[width=\textwidth]{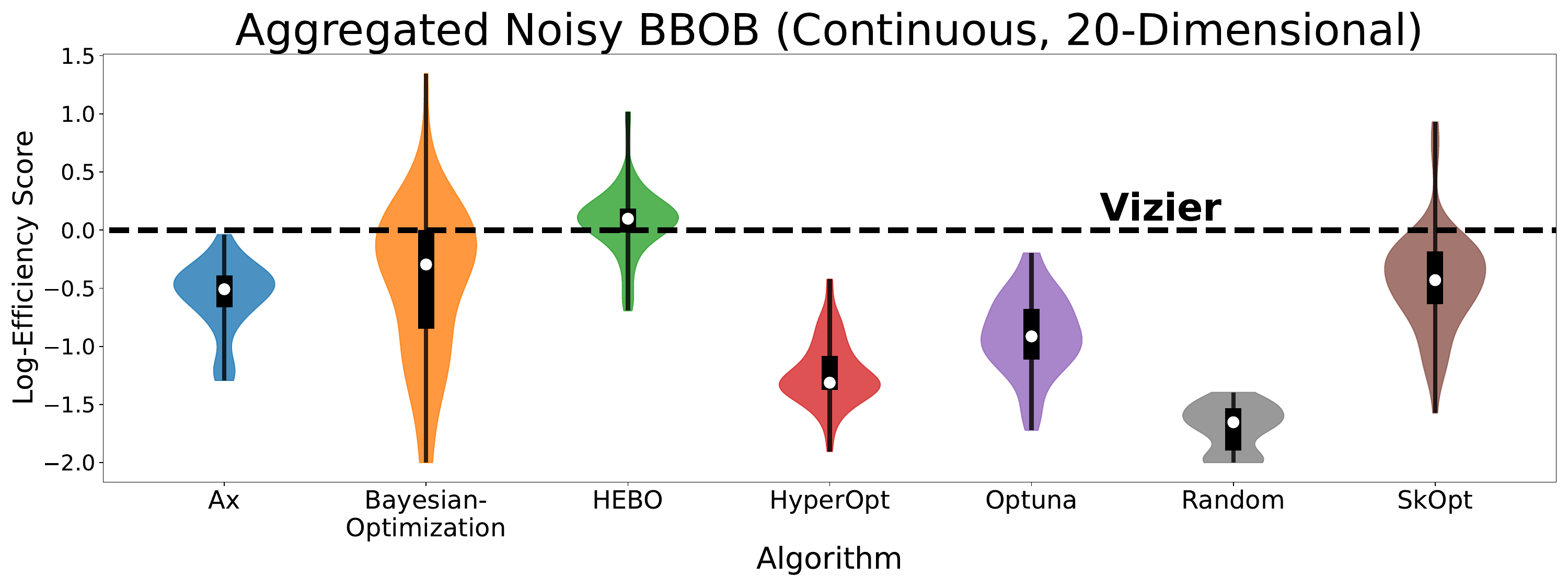}
    \caption{Higher is better $(\uparrow)$. Violin plots displaying distributions of log-efficiency scores across the cartesian product of all BBOB functions and noise models.\vspace{0.2cm}}
    \label{fig:noisy_continuous_20d_violin}
\end{figure}

In Figure \ref{fig:noisy_continuous_20d_violin}, we benchmark over all three noise models (with their strongest suggested ``severe'' settings) over every BBOB function, and see that the plot is similar to the original Figure \ref{fig:continuous_20d_violin}, albeit with a few additional positive log-efficiency outliers from baselines.

We hypothesize that since all of the baselines preprocess and normalize the observed $y$-values, the rankings of trials remain fairly stable, and thus the baseline performances remain robust to noise. Furthermore, small differences in the optimization trajectory generally do not affect the log-efficiency metric, which holistically measures the convergence curve.

\vspace{0.2cm}

\clearpage

\section{Vizier GP Bandit: Extended Details}
\label{appendix:vizier_gp_bandit_extended}
This section serves to clarify additional details for readers who are less familiar with Bayesian Optimization or are interested in lower-level specifics.

\subsection{Kernel Definition}
We use the Matern-5/2 kernel, whose exact implementation is found in TF Probability\footnote{\url{https://www.tensorflow.org/probability/api_docs/python/tfp/math/psd_kernels/MaternFiveHalves}}. Naturally, for continuous parameters, we may define the distance between two features $\vecx_{i}$ and $\vecx_{j}$ to be Euclidean-based. However, different coordinates may have different scales, and thus we need to account for normalization. We thus compute a \textit{length-scaled} distance, defined as:
\begin{equation} 
\delta(\vecx_{i}, \vecx_{j} )^{2} = 5 \cdot \sum_{d=1}^{D} \frac{\left(\vecx_{i}^{(d)} - \vecx_{j}^{(d)} \right)^2}{\lambda^{(d)}}
\end{equation}
where $\vec{\lambda} = \left(\lambda^{(1)}, \ldots, \lambda^{(D)} \right) =  \exp (\veclambdalog)$ are the squared length-scales. For \texttt{CATEGORICAL} parameters, the contribution from each parameter $x^{(c)}$ is instead \begin{equation}\frac{\mathbbm{1}\left(x^{(c)}_{i} \neq x^{(c)}_{j}\right)^2}{\lambda^{(c)}} \end{equation} where $\lambda^{(c)}$ is a single trainable value. This logic is specifically implemented by the\\ \texttt{FeatureScaledWithCategorical} kernel in TF Probability.

Using $\delta = \delta(\vecx_{i}, \vecx_{j})$ for brevity, this distance will then be used to define the kernel:
\begin{equation}
K(\vecx_{i}, \vecx_{j}) = \alpha^{2} \cdot \left(1 + \delta + \frac{\delta^{2}}{3}\right) \cdot  \exp\left(-\delta \right)
\end{equation}
where $\alpha = \exp(\alpha_{\log}) \in \R$ is an additional amplitude argument. This kernel together with a zero prior mean thus defines the Gaussian Process.

\clearpage

\subsection{Firefly Acquisition Optimizer}
For completeness, we formalize the vectorized firefly mechanic in Algorithm \ref{alg:modified_firefly}.

\vspace{0.3cm}
\begin{algorithm}[h]
\caption{Modified Firefly Algorithm}\label{alg:modified_firefly}
\begin{algorithmic}[1]
\State \textbf{Input:} Pool size $\poolsize \in \mathbb{N}$, Batch size $\flybatchsize$ dividing $\poolsize$, \texttt{MAX\_ITERATIONS}, acquisition function $a(\cdot)$
\State \textbf{Output:} Feature vector $\vecx$ 
\State \textbf{Goal:} Find $\argmax \set{a(\vecx) : \vecx \text{ is feasible}}$.

\\
\State Initialize firefly pool $\mathbf{X} \in \R^{\poolsize \times D}$ with $\poolsize$ random feasible features.
\State Partition the firefly pool into batches $\set{\flybatch_i \in \R^{\flybatchsize \times D} : i = 1, 2, \ldots, \poolsize / \flybatchsize}$ of size $\flybatchsize$.  
\For{iteration $\in \{1, \ldots, \texttt{MAX\_ITERATIONS} \}$ }
\For{$i \in \set{1, 2, \ldots, \poolsize / \flybatchsize }$}
\State Update $i$-th batch features $\flybatch_{i}$ using forces from the entire pool, per Eq. \ref{eq:vectorized_firefly}.
\State Evaluate all features vectors in $\flybatch_{i}$ on $a(\cdot)$ and persist their scores.
\State Retain the feature vector $\vecx_\text{best}$ with the maximum score seen throughout.
\EndFor
\EndFor
\State Return $\vecx_\text{best}$, the highest scoring feature vector of all those evaluated.
\end{algorithmic}
\end{algorithm}

\subsection{Batched Optimization}
We formalize the pure exploration batched method in Algorithm \ref{alg:gp_ucb_pe}.

\vspace{0.3cm}
\begin{algorithm}[h]
\caption{GP-UCB-PE for Batched Suggestions at Iteration $t$}
\begin{algorithmic}
\State{\textbf{Input}:
\begin{itemize}
\item $\mathcal{D}$: Set of all evaluated trials and their function values.
\item $\mathcal{U}$: Set of all unevaluated trials with dummy function values of zeros.
\item $\beta \geq 0$ and $\beta_e \geq 0$: UCB parameters.
\item $\rho > 0$: Penalty parameter for points outside of the promising region.
\item $0 \leq q < 1$: Probability for overwriting the UCB selection with PE.
\item $B \geq 1$: Number of suggestions to generate in a batch.
\end{itemize}
}
\For{$b \in \{1, \ldots, B\}$}
\State {
\vspace{-1.5em}
\begin{eqnarray}
x^* &\leftarrow& \argmax_{x \in \mathcal{D} \cup \mathcal{U}} \> \> \mbox{UCB}(x | \mathcal{D}, \emptyset, \beta), \mbox{ where } \emptyset \mbox{ denotes the empty set.} \\
\tau &\leftarrow& \mu(x^* | \mathcal{D}).\\
\Delta &\leftarrow& \mathbbm{1}(\mathcal{D} \mbox{ contains trials evaluated after the generation of the latest trial in } \mathcal{U}). \\
x_b &\leftarrow& 
\begin{cases}
\arg \max_{x \in \X} \mbox{UCB}(x | \mathcal{D}, \mathcal{U}, \beta), & \mbox{with prob. $1-q$ if $\Delta$ is true (c.f., Eqn~\ref{eqn:ucb-def}).} \\
\arg \max_{x \in \X} \mbox{PE}(x |  \mathcal{D}, \mathcal{U}, \tau, \beta_e, \rho), & \mbox{otherwise (c.f., Eqn~\ref{eqn:pe-def}).}
\end{cases} \\
\mathcal{U} &\leftarrow& \mathcal{U} \cup \{(x_b, 0)\}.
\end{eqnarray}
}
\EndFor
\State{\textbf{Output:} $\{x_1, x_2, \ldots, x_B\}$.} 
\end{algorithmic}
\label{alg:gp_ucb_pe}
\end{algorithm}

\subsection{Just-in-time (JIT) Compilation Optimization}
JAX \citep{jax} utilizes just-in-time compilation to optimize a given computation graph. Compiled graphs are stored into a global cache, which may only be used if tensor input/output shapes are fixed, and expensive re-compilation must occur if new shapes are encountered. 

This is an issue for Bayesian optimization, where multiple components depend on the number of observed trials $t$, such as the Gaussian process's kernel matrix which is of shape $(t \times t)$. Furthermore, if the algorithm is to be re-used for different search spaces, the feature dimension $D$ may also change.

To reduce the cost of re-compilation, we reduce the number of unique encountered shapes, by \textit{padding} the collection of inputs $\left[ \vecx_{1}, \ldots, \vecx_{t} \right]^\top \in \mathbb{R}^{t, D}$ to fixed dimensions according to some schedule, e.g., powers of $2$. 

\subsection{Exact Hyperparameters} \label{sec:exact-hps}
All exact hyperparameters can be observed from the code\footnote{\url{https://github.com/google/vizier/blob/main/vizier/_src/algorithms/designers/gp_ucb_pe.py} and \url{https://github.com/google/vizier/blob/main/vizier/_src/algorithms/designers/gp_bandit.py}} in \texttt{google-vizier[jax]==0.1.7}.

\textbf{MAP Estimation:} To optimize the kernel hyperparameters $\alpha_{\log}, \veclambdalog, \varepsilon_{\log}$, our L-BFGS-B uses 50 maximum iterations, each step using a maximum of 20 line search steps. The optimizer is restarted for 4 times each with different random initializations, eventually returning the best hyperparameters.

\textbf{Trust Region Schedule:} The radius is scheduled based on the number of observed points $t$ and post-processed dimension $D$, specifically:
\begin{equation}
\text{radius} = 0.2 + (0.5 - 0.2) \cdot \frac{1}{5}\frac{t}{(D+1)}
\end{equation}
If this radius is $>0.5$, then the trust region is disabled (i.e. the radius becomes infinite).

\textbf{Firefly Optimizer:} We use a maximum of $75000$ evaluations, with a batch size of $\flybatchsize =25$ and pool size $P = \min\{10 + \frac{1}{2}D + D^{1.2}, 100 \}$, thus leading to a maximum of $\texttt{MAX\_ITERATIONS} = \floor{75000 / P}$. Further hyperparameters below control the specific update rules. 

For force computations, $\gamma = 4.5 / D$, $\eta_{\text{attract}} = 1.5$, and $\eta_{\text{repel}} = 0.008$.

For perturbations, if the space is hybrid, then $\omega_{\text{continuous}} = 0.16$ and $\omega_{\text{categorical}} = 1.0$, whereas if the space is purely categorical, $\omega_{\text{categorical}} = 30$ for better exploration. Unsuccessful fireflies (i.e. those which did not improve acquisition score have their perturbation further scaled down by $0.7$.

The pool of fireflies also has a \textit{keep probability} of 0.96, in which new random fireflies may be introduced if a firefly is not kept.


\textbf{Multi-objective Acquisition:} We use 1000 scalarizations for hypervolume approximation, and define our hypervolume reference point \citep{refpoint} as $\y_{\text{ref}} := \vecy_{\text{worst}} - 0.01 \cdot (\vecy_{\text{best}} - \vecy_{\text{worst}})$, where $\vecy_{\text{best}}$ and $\vecy_{\text{worst}}$ are, respectively, the maximum and minimum values of post-processed metrics.

\textbf{Batched Optimization:} We maintain $\sqrt{\beta} = 1.8$ and set the exploration-specific UCB parameter $\sqrt{\beta_{e}} = 0.5$. We further set the penalty $\rho = 10.0$ and overwrite probability $q = 0.1$. 

\textbf{Quasi-Random Seeding:} For single-objective problems, we did not use initial quasi-random trial sampling. For multi-objective problems, the initial 10 trials are quasi-randomly sampled.





\clearpage

\section{Experimental Baselines}
\label{appendix:baselines}
For every baseline and benchmark function, we ran the optimization loop with 20 repeats, with a horizon of 100 trials.

\subsection{Packages}
At the time of writing, we used Python package \texttt{ray[default]==2.10.0} along with the following baseline packages:
\begin{itemize}
\item \texttt{ax-platform==0.3.4} \citep{botorch_official}
\item \texttt{bayesian-optimization==1.4.3} \citep{bayesian_optimization_github}
\item \texttt{HEBO==0.3.5} \citep{hebo}
\item \texttt{hyperopt==0.2.7} \citep{hyepropt}
\item \texttt{optuna==3.6.1} \citep{optuna}
\item \texttt{scikit-optimize==0.10.1} \citep{scikit_optimize}
\end{itemize}
Vizier used \texttt{google-vizier[jax]==0.1.7}. 

Since Ray Tune does not support multi-objective optimization, we directly wrapped the baseline packages (\texttt{ax-platform}, \texttt{HEBO}, \texttt{optuna}) into OSS Vizier's API for comparison.

\subsection{Algorithm Summaries}
\label{appendix:baseline_details}
Below, we summarize the key components influencing each baseline's behavior. 
All GP-based methods use a Matern kernel prior and L-BFGS-B for MAP estimation. The major difference among them are the choice of acquisition function and corresponding acquisition optimizer.

\textbf{Ax/BoTorch:} Depending on search space, \texttt{choose\_generation\_strategy}\footnote{\url{https://github.com/facebook/Ax/blob/main/ax/modelbridge/dispatch_utils.py}} chooses the algorithm. The number of continuous vs. categorical parameters, affects the choice of GP model \citep{saasbo_pytorch} and kernel. The default acquisition is an expected improvement (EI)-based \texttt{qNoisyExpectedImprovement}. Note we do not provide objective noise variance in advance. For acquisition optimization, L-BFGS-B is used for continuous spaces while a sequential greedy algorithm is used for mixed and categorical spaces.

\textbf{BayesianOptimization:} Uses the default sklearn \texttt{GaussianProcessRegressor} with a Matern kernel (with slightly different hyperparameters), with UCB ($\sqrt{\beta} = 2.576$). Acquisition optimization is L-BFGS-B with an initial random search warmup\footnote{  \url{https://github.com/bayesian-optimization/BayesianOptimization/tree/master/bayes_opt}}.

\textbf{HEBO\footnote{\url{https://github.com/huawei-noah/HEBO/blob/master/HEBO/hebo/optimizers/hebo.py}}:} Uses the \texttt{MACE} acquisition, which is a maximum over expected improvement (EI), probability of improvement (PI), and upper confidence bound (UCB). The acquisition is optimized by NSGA-II. Note: Empirically, the algorithm tended to stop early on multiple problems due to numerical instabilities. In such cases, we extended the best-so-far curve with the latest value to properly compute log-efficiency comparison metrics.

\textbf{HyperOpt\footnote{\url{https://github.com/hyperopt/hyperopt/blob/master/hyperopt/tpe.py}}:} Uses Tree-Structured Parzen Estimator (TPE) \citep{tpe}
An adaptive variant is available\footnote{\url{https://github.com/hyperopt/hyperopt/blob/master/hyperopt/atpe.py}} as well.

\textbf{Optuna:} This package wraps multiple algorithms, but by default also uses TPE\footnote{\url{https://optuna.readthedocs.io/en/stable/reference/samplers/generated/optuna.samplers.TPESampler.html}}, along with its multi-objective variant\footnote{\url{https://optuna.readthedocs.io/en/v2.10.1/reference/generated/optuna.samplers.MOTPESampler.html}} (MOTPE) \citep{motpe}.

\textbf{Scikit-Optimize (SkOpt):} See the \texttt{Optimizer} class\footnote{\url{https://github.com/scikit-optimize/scikit-optimize/blob/master/skopt/optimizer/optimizer.py}}. The default regressor is a GP with a Matern kernel for continuous and hybrid spaces, and \texttt{HammingKernel} for categorical-only spaces, along with L-BFGS-B and random search as acquisition optimizers, respectively. The acquisition is \texttt{``gp\_hedge"}, which randomly chooses an acquisition among lower confidence bound (LCB), expected improvement (EI), and probability of improvement (PI).

\subsection{Ablations}
\textbf{L-BFGS-B Latency:} In Figure \ref{fig:eagle_vs_lbfgsb_acquisition}, we could not fairly measure the duration of L-BFGS-B directly, as its \texttt{scipy} implementation is not fully JAX-optimized, and also early-stops. In comparison, Firefly is fully JAX-optimized and runs for the entire iteration count. To make the timing comparison fair, we instead use a lower bound for the duration of a hypothetical L-BFGS-B implementation that had been highly JAX-optimized and did not use early-stopping. Our lower bound is: \begin{equation}\text{(duration of gradient computation in JAX)} \times \text{(maximum iterations)}\end{equation} which assumes that the actual algorithm and its line-search steps (which could call more gradient computations) are instantaneous.

\textbf{L-BFGS-B in Bayesian Optimization Loop:} In Figure \ref{fig:e2e_bayesopt_eagle_vs_lbfgsb}, our settings for L-BFGS-B are maximized to ensure convergence, with 100 maximum iterations and 25 random restarts.

\clearpage

\section{Benchmarks}
\label{appendix:benchmark_objectives}
\subsection{BBOB}
We use the standard Black-Box Optimization Benchmark \citep{bbob} set. Note that in this paper, $f(x)$ is an objective to be maximized, and thus we negate the standard BBOB functions (which are designed for minimization benchmarks).

\textbf{Randomized Shifting:} Since the global optimum of every BBOB function is defined at the origin and the every parameter range is $[-5, 5]$, the initial centering trial would trivialize this benchmark. Therefore, at every repeat study, we randomly shift a base BBOB function $f$ to $f(x-c)$, where each coordinate of $c$ is uniformly sampled from $[-5,5]$. Note that every coordinate value of $x-c$ is now within $[-10, 0]$ and can be outside of the original search space, but can still be evaluated as BBOB functions naturally have an unrestricted domain.

\textbf{Categorization:} To convert a \texttt{DOUBLE} parameter configuration to a \texttt{CATEGORICAL} configuration, we simply select 10 equidistant grid points from $[-5,5]$ as the categories.

\textbf{Noise:} The noise models (with ``severe'' settings) from \cite{noisy_bbob} wrap a deterministic $f(x)$ and return a distribution as follows:
\begin{itemize}
\item Gaussian Noise (Multiplicative): $f(x) \cdot \exp \left(\mathcal{N}(0,1)\right)$
\item Uniform Noise (Multiplicative): $f(x) \cdot U(0,1) \cdot \max \left(1, \left(\frac{10^{9}}{-f(x) + \epsilon} \right) \right)^{(0.49 + 1/D) \cdot U(0,1)}$
\item Cauchy Noise (Additive): $f(x) - \max \left(0, 1000 + \mathbbm{1}_{U(0,1) < 0.2} \frac{\mathcal{N}(0,1)}{\abs{\mathcal{N}(0,1)} + \epsilon} \right)$
\end{itemize}
Here, we set $\epsilon=10^{-199}$ to avoid division-by-zero, $U(0,1)$ is the uniform distribution over $[0,1]$, and $\mathcal{N}(0,1)$ is the standard normal. Note that in expressions with multiple distributions we sample independently for each.

\subsection{COMBO}
We use the functions introduced in \citep{combo}. Unlike BBOB functions, these functions are ``purely'' categorical, i.e., their implementations are defined purely in terms of the categories and are not wrapped variants of continuous space objectives.

\textbf{Permuted Feasible Values:} Similar to the centering issue with BBOBs, an initial centering trial (i.e. choosing the middle index across the list of categories) can immediately lead to a large advantage. Therefore at every repeat study, we randomly permute each list of categories. For example, suppose $\X = C_1 \times C_2 \times ... \times C_d$ where each $C_i$ is a set of categories. Instead of evaluating $f(x)$, we would instead fix random permutations $\pi_i$ over $C_i$ at the start of the study, and then evaluate 
$f_{\pi}(x) := f(\pi_1(x^{(1)}), \pi_2(x^{(2)}), \ldots, \pi_d(x^{(d)}))$ as our benchmark objective. 

\subsection{Multi-Objective}
We used the \texttt{optproblems} package, which implements the benchmark family and problem IDs: DTLZ (1-7) \citep{dtlz_multiobjective}, WFG (1-9) \citep{wfg_multiobjective}, and ZDT (1-4, 6) \citep{zdt_multiobjective} multi-objective benchmarks. ZDT5 was not used as it does not allow variable dimensions. For WFG, we set its number of ``position-related parameters" to $M-1$.

\textbf{Metric Normalization:} We encountered varying magnitudes across different metrics across the \texttt{optproblems} functions. Having different scales can adversely affect hypervolume prioritizations (e.g. algorithms may focus only on a single objective as it contributes to most of the hypervolume). We thus first normalize each objective $f^{(i)}(x)$ by dividing by $\frac{1}{\abs{\mathcal{G} }}\sum_{x \in \mathcal{G}} \abs{f^{(i)}(x)}$ where $\mathcal{G}$ consists of 100 evenly-spaced grid points along the search space diagonal.

\textbf{Hypervolume Computation:} As mentioned in Section \ref{subsec:multiobjective}, computing exact dominated hypervolumes is \textbf{\#P}-hard and thus we instead used the scalarization method from \cite{golovin2020random}, with 10000 scalarization weights, which sufficiently approximates the hypervolume within $10^{-4}$ error. Our reference point was created from the worst metrics among all $y$-values from all algorithms.

\clearpage

\section{Evaluation Protocols}
\label{appendix:evaluation_protocol}
\subsection{Log-Efficiency (Extended)}
There is a plethora of previous works that examine the complex question of benchmarking and comparing optimization algorithms. As noted in \cite{hansen2022anytime} on the topic of anytime benchmarking, both ranking-based metrics and objective-based metrics can be biased as information on the improvement magnitudes can be lost or warped. They argue that convergence runtime is the only performance measure with a generic, meaningful, and quantitative interpretation. 

Thus, in order to aggregate performances across different benchmark objectives with vastly different magnitudes of $y$, for each objective function, we compute the \textit{log-efficiency} score, which is a variant of performance profiles and measure relative convergence speed. The notion of log-efficiency can be derived from the popular work on \textit{performance profiles} \citep{dolan2002benchmarking} which indeed uses convergence runtime ratios and generalizes the idea for multiple algorithms. 
Unfortunately, the notion is vaguely defined - i.e., the \emph{required budget} in this case is defined to be the computing time or iterations required to ``solve" a benchmark $\mathcal{B}$ using algorithm $\mathcal{A}$. The performance ratio is then defined to be 
\begin{equation}
\frac{\rbudget{\mathcal{B} | \alg}}{\rbudget{\mathcal{B} | \alg_{best}}} 
\end{equation}
Note that this is at least $1$ and can be max-clipped when too large. The performance profile is the distribution of performance ratios, assuming the benchmark is uniformly drawn at random.

While the performance profile provides meaningful normalized metrics across different benchmark problems in $\mathcal{B}$, it requires comparison to the best overall algorithm and is vaguely defined. For black-box optimization, as mentioned in the main body, we define the required budget as follows.

\begin{definition}[Required Budget]
The \emph{required budget} is the expected number of iterations needed to reach a target $y$.
For a fixed objective $f$ and algorithm $\alg$ generating a random trajectory $\set{(x_t, y_t) : t \ge 0}$ when run on $f$:
\[
\rbudget{y \ | \ f, \alg} := \expct{\min \set{t : y_t \ge y}}
\]
\end{definition}
We define \emph{efficiency ratio} analogously, 
\[
\eratio{y | f, \alg, \alg'} := \frac{\rbudget{y | f, \alg}}{\rbudget{y | f, \alg'}}
\]
We may then take the logarithm of our efficiency ratio to make it zero-centered and symmetric:
\begin{equation}
    \text{LogEfficiency}(y \> | \> f, \mathcal{A}) := \log(\eratio{y | f,  \operatorname{Vizier}, \alg}) = \log\left(\frac{\rbudget{y \> | \>  f, \text{Vizier}}}{\rbudget{y \> | \> f, \mathcal{A} }} \right)
\end{equation}
Note that a positive log efficiency score means the compared algorithm uses less iterations, by a factor of $\exp(-\text{LogEfficiency})$, to reach the same objective than Vizier. We clip the LogEfficiency to be in $[-2, 2]$ to avoid outliers. 

The choice of how to set $y$ or the sequence of $y$'s is important and nuanced. As an example, if we simply define the sequence of $y$'s as Vizier's best-so-far curve, the log efficiency scores may be uninformative when Vizier is significantly faster or slower than the compared curve. We set $y$ to be the averaged curve between $\mathcal{A}$ and Vizier's best-so-far curves, and allows us to take the median of these log-efficiencies as our final representative score.

Formally, suppose $\set{(x_t(\alg), t_t(\alg)) : t \ge 0}$ is a random trajectory when running $\alg$ on $f$, and let $z_t(\alg) := \expct{\max_{\le t} x_t(\alg)}$ be the expected best performance seen within the first $t$ trials.
Since we cannot compute the expectations exactly, we use the empirical mean over sampled trajectories.
Let 
$\overline{z}(\alg, \alg') := \set{\frac{1}{2}\paren{z_t(\alg) + z_t(\alg')} : t = 1, 2, \ldots, T}$
where $T$ is the overall trial budget.
To compare Vizier with $\alg$, we report the median of 
\[
\set{ \frac{\rbudget{y | f, \operatorname{Vizier}}}{\rbudget{y | f, \alg}} \> \> \middle\vert \> \> y \in \overline{z}(\alg, \operatorname{Vizier}) }.
\]

\end{document}